%% file: arxiv.tex
\documentclass{article}

\PassOptionsToPackage{numbers, compress}{natbib}
\usepackage[preprint]{neurips_2026}
\makeatletter
\renewcommand{\@noticestring}{Accepted at the 40th Conference on Neural Information Processing Systems (NeurIPS 2026).}
\makeatother
\usepackage{amssymb}

\usepackage[utf8]{inputenc} 
\usepackage[T1]{fontenc}    
\DeclareFontShape{T1}{ptm}{m}{scit}{<->ptmrc8t}{}
\usepackage{hyperref}       
\usepackage{url}            
\usepackage{booktabs}       
\usepackage{amsfonts}       
\usepackage{nicefrac}       
\usepackage{silence}
\usepackage{microtype}      
\PassOptionsToPackage{table}{xcolor}
\usepackage{xcolor}         
\usepackage{algorithm}
\usepackage{algpseudocode}
\usepackage{multirow}
\usepackage[table]{xcolor}
\usepackage{array}
\usepackage{fontawesome5}

\usepackage{mdframed}

\usepackage{tikz}
\usetikzlibrary{positioning}

\usepackage{float}
\usepackage{wrapfig}
\usepackage{booktabs}
\usepackage{tabularx}
\usepackage{longtable}
\usepackage{array}
\usepackage{xcolor}

\title{\method{}: Continuous Speculation for Accelerating Multi-Hop Retrieval Agents}

\author{%
  Mehrdad Saberi\thanks{Equal contribution} \quad
  Keivan Rezaei\footnotemark[1]\hspace{0.45em}\thanks{Now at Google Research.} \quad
  Soheil Feizi \\
  University of Maryland, College Park \\
  \texttt{\{msaberi, krezaei, sfeizi\}@umd.edu}\\
  \\
  \faGithub~\textbf{Project}: \url{https://github.com/mehrdadsaberi/spechop}
}

\usepackage{amsmath}
\newtheorem{theorem}{Theorem}
\newtheorem{assumption}{Assumption}
\newtheorem{corollary}[theorem]{Corollary}
\newtheorem{observationth}{Observation}

\newcommand{\method}{\textsc{SpecHop}}
\newcommand{\state}{\mathbf{s}}
\newcommand{\action}{\mathbf{a}}
\newcommand{\observation}{\mathbf{o}}
\newcommand{\seg}{\boldsymbol{\tau}}

\newcommand{\Tseg}{T_{\mathrm{seg}}}
\newcommand{\Ttool}{T_{\mathrm{target}}}
\newcommand{\Thop}{T_{\mathrm{hop}}}
\newcommand{\Tspec}{T_{\mathrm{spec}}}

\newcommand{\Thopspec}{T_{\mathrm{hop}}^{\mathrm{spec}}}

\newcommand{\thread}[1]{$\textsc{T}_{#1}$}
\newcommand{\Latency}{\mathcal{L}}

\newcommand{\generatormodel}{\mathcal{M}}
\newcommand{\speculator}{\mathcal{S}}
\newcommand{\targettool}{\mathcal{T}}
\newcommand{\verifier}{\mathcal{V}}
\newcommand{\RelLatency}{\mathrm{RelLat}}
\newcommand{\RelLatencylive}{\RelLatency_{\mathrm{live}}}

\newcommand{\subq}{sub-question}
\newcommand{\suba}{sub-answer}

\newcommand{\expected}[1]{\mathbb{E}\left[#1\right]}

\begin{document}
\maketitle

\begin{abstract}
Large language models increasingly use external tools such as web search and document retrieval to solve information-intensive tasks.
However, multi-hop tool use in complex tasks introduces substantial latency, since the model must repeatedly wait for tool observations before continuing.
We study how to accelerate such trajectories without changing the final trajectory the model would have taken without acceleration, assuming access to faster but less reliable speculator tools.
We develop a theoretical framework for lossless speculation in multi-hop tool-use settings, characterizing the optimal achievable latency gain.
We propose \method{}, a continuous speculation framework that maintains multiple speculative threads, verifies predicted observations asynchronously as target tool outputs arrive, commits correct branches, and rolls back incorrect ones.
This preserves accuracy while reducing wall-clock latency. We show that \method{} can approach oracle latency gains with enough active threads. Empirically, on retrieval-augmented multi-hop tasks, \method{} closely matches theoretical predictions and reduces latency by up to 40\% in some settings.
\end{abstract}

\section{Introduction}

Large language models (LLMs) are increasingly equipped with external tools, including web search, document retrieval, code execution environments, and APIs~\cite{schick2023toolformer,yao2022react,openai_tools_docs,google_gemini_function_calling,anthropic_tool_use_docs}.
This enables more complex tasks, such as deep research, agentic workflows, interactive decision making, and information-intensive reasoning that requires interaction with external environments~\cite{wang2023surveyagents,wang2023voyager,openai_deep_research,google_deep_research,wu2026deepresearch}.
In these settings, the model's trajectory is no longer a single forward generation, but a \textit{multi-step} process of reasoning, tool calls, observations, and further reasoning.
In this work, we focus on information-gathering tools, such as web search and database retrieval~\cite{yang2018hotpotqa,ho2020constructing,trivedi2022musique,wang2025corag}.

Enabling language models to call external tools improves performance, but also introduces substantial execution latency~\cite{ye2025speculative,sui2026act}.
Because inference trajectories are sequential, the model may reason, wait for a web search result, reason again, retrieve documents from a database, and continue this process over multiple hops.
Thus, total latency depends not only on model decoding, but also on external tool calls.
Recent profiling studies show that this bottleneck is significant in practice: in deep-research agents, web search can account for 73\% of end-to-end latency on average for GPT-5, and up to 91\% in more complex cases~\cite{abhyankar2025demystifying}.
In this work, we accelerate such \textit{multi-hop} tool-use trajectories without changing the trajectory the model would have taken without acceleration.

\citet{ye2025speculative} propose \emph{Speculative Actions}, which assume access to a speculator that predicts a tool call return with non-zero success probability $p$.
They show that such speculators can predict target tool outcomes with lower latency in practice.
Their method speculates one hop while the original tool call runs in parallel, then verifies the result to preserve accuracy and reduce latency when speculation succeeds.
Building on this setup, we develop a theoretical framework for lossless speculation in \textit{multi-hop} tool-use trajectories.
We first study an oracle-verification setting, where speculative correctness is known immediately, yielding an upper bound on the latency improvement achievable by any lossless speculative method.

Motivated by our theoretical analysis of such setup,
we propose \method{}, a continuous speculation framework that pushes toward the oracle latency gain.
Rather than waiting idly for each target tool call to return, \method{} \textit{keeps} a speculative pipeline active throughout the trajectory.
It maintains \textit{multiple parallel threads}, each corresponding to a different speculative stage of the same underlying trajectory, and \textit{continues speculating future hops} while previous target calls are still running.
New threads are created only while the earliest target tool call has not yet returned, so a slower target tool admits more threads and a larger latency reduction (Section~\ref{sec:cont-spec}).
Those threads wait on the target tool rather than on the model: in our runs, concurrent model calls stay about 2 instead of growing with the number of threads (Appendix~\ref{app:live}), and the resulting per-call times increase only slightly (Table~\ref{tab:live-contention}).
When a target output becomes available, a verifier checks whether the corresponding speculative observation is equivalent to the target observation.
If verification succeeds, \method{} commits the speculative branch and extends it with new speculative threads; if verification fails, it discards downstream speculative work and resumes from the last verified state (See Figure~\ref{fig:specpipe-overview}).
Thus, \method{} preserves the original trajectory under verification while reducing wall-clock latency.


We theoretically study \method{} and show that it can close the gap to the oracle latency gain while remaining lossless.
We characterize the required number of active threads in terms of the speculator's relative latency and the fraction of time spent on external tool calls.
Empirically, we evaluate \method{} on retrieval-augmented multi-hop QA tasks (2WikiMultihopQA~\cite{ho2020constructing}, MuSiQue~\cite{trivedi2022musique}, and DeepResearch-9K~\cite{wu2026deepresearch}) with web search or Wikipedia retrieval as the external tool, and either an LM-based speculator or a fast cache-based speculator.
Across these settings, latency matches our theoretical predictions, with \method{} preserving task accuracy while achieving up to $40\%$ latency gain over standard multi-hop execution.

\input{figures/teaser}

\section{Related Work}

\paragraph{Speculative decoding.}
Autoregressive token generation in large language models can create memory-bandwidth bottlenecks during inference~\cite{leviathan2023fast,yan2025decoding}.
Speculative decoding mitigates this through a ``draft-then-verify'' paradigm, where a smaller draft model proposes tokens that are verified by a larger target model, yielding mathematically lossless outputs that match the target model distribution~\cite{chen2023accelerating,leviathan2023fast}.
Recent work extends this idea with multiple drafters~\cite{mcdanel2025pipespec}, parallel speculative pipelines~\cite{yin2025specpipe}, and draft/target models with different vocabulary spaces~\cite{timor2025accelerating}.
Our work shares the draft-then-verify philosophy, but speculates over tool observations rather than tokens.

\paragraph{Foundations of speculative execution.}
Speculative execution originates in computer architecture, where it is used to overcome instruction-level bottlenecks and avoid stalls from conditional branches~\cite{smith1981study}.
Branch-prediction methods estimate future control flow using history registers and execution patterns~\cite{yeh1991two}, while speculative mechanisms execute predicted instructions before they are guaranteed to be needed, verify them once the true control flow is known, and roll back upon misprediction~\cite{smith1988implementing,tomasulo1967efficient}.
\method{} adapts this classical systems principle from instruction-level execution to trajectory-level tool use in language model agents.

\paragraph{Agentic tool use and retrieval-augmented generation.}
Equipping language models with external tools, especially retrieval~\cite{lewis2020retrieval} and search, has become central to retrieval-augmented generation and agentic frameworks~\cite{radhakrishnan2024knowing, wang2025deepnote,singh2025agentic,yang2026sparc}.
Accelerating retrieval-augmented agents has recently been studied from several complementary angles.
\citet{wang2024speculative} focus on static retrieval at the beginning of the pipeline, using multiple speculative RAG models with different retrieved documents and later verifying with a larger generalist LLM to reduce latency without sacrificing accuracy.
\citet{sui2026act} propose anticipating tool-call needs earlier by identifying patterns in the model's generation, allowing the system to launch tools before the model explicitly reaches the tool-call point.
Closest to our work, \citet{ye2025speculative} show that some external calls can be replaced by faster speculative models, providing empirical evidence that individual tool-use hops can sometimes be accelerated through speculation.
Their procedure speculates one hop ahead and then waits. Section~\ref{sec:results} compares it with continuous \method{} on the same traces.

\section{Accelerating Multihop Tool-Use Trajectories}

Language models capable of tool use answer queries by invoking external tools when additional information or computation is needed.
They often decompose a query into intermediate steps, each resolved either from parametric knowledge or through tools such as web search, document retrieval, or databases of documents such as Wikipedia articles.
This behavior is common in settings studied by DeepResearch~\cite{wu2026deepresearch} and multihop question-answering datasets~\cite{ho2020constructing,trivedi2022musique,yang2018hotpotqa}.
Given a query $q$, the model $\generatormodel$ follows an iterative tool-augmented trajectory: it generates intermediate reasoning, calls tools, receives observations, updates its state, and repeats this process until producing the final answer.

Formally, for a query $q$, the model $\generatormodel$ follows an inference trajectory over states
$\state_0,\state_1,\dots,\state_N$, where $\state_0$ is induced by the query and $\state_N$ is terminal.
At each state $\state_i$ for $0 \leq i < N$, the model conditions on the prior context, generates an intermediate segment $\seg_i$, and selects a tool action $\action_i$.
The tool returns an observation $\observation_i$, which is added to the context to form the next state $\state_{i+1}$.
From $\state_N$, the model generates the final answer $f$.

\input{figures/state_machine}

Thus, the trajectory consists of $N$ hops.
Because the model typically waits for each tool response before proceeding, inference latency includes both language-model decoding time and external-tool latency from web search, retrieval, code execution, or database queries.


\paragraph{System latency.}
Processing a query $q$ requires the model to complete all $N$ hops: generating intermediate segments $\seg_i$, selecting tool actions $\action_i$, receiving observations $\observation_i$, and transitioning to the next state.
We model the wall-clock time of each component as a random variable.
Let $\Tseg$ be the time to generate an intermediate segment, and let $\Ttool$ be the time for the target tool to return its result and for the model to consume the observation before continuing.
Thus, the latency of one sequential hop is
\[
\Thop = \Tseg + \Ttool .
\]
For a query requiring $N$ hops, the expected total time to reach the final state is $N\expected{\Thop}$.

\subsection{Access to a Speculator Tool}

Following~\cite{ye2025speculative}, we assume access to a speculator tool that can replace the original target tool.
The speculator has two key properties.
First, unlike the target tool, the speculator is approximate: there is uncertainty about whether its returned observation is true.
Second, the speculator is faster than the target tool, and therefore can potentially reduce wall-clock latency when its output is correct.

\paragraph{Speculator accuracy.}
The speculator's accuracy depends on the target tool and task.
Following \citet{ye2025speculative}, we model the speculator as succeeding at each hop with probability $p>0$, where success means that the speculative observation leads the model to the same next state as the target-tool observation.
This can occur when the needed factual information is already captured by an off-the-shelf model's parametric knowledge, or is available in a smaller database or cache.
Prior work suggests such speculators can be useful in practice, achieving nontrivial success probabilities.

\paragraph{Speculator latency.}
Let $\Tspec$ be the random variable denoting the latency of the speculator, including the time required to produce the speculative observation and enable the model to transition to the next state.
We assume that the speculator is more efficient than the target tool, i.e., $\Tspec < \Ttool$.

We make the following assumptions to leverage speculators while preserving the target-tool trajectory.

\begin{assumption}[Fast approximate speculator]
\label{assump:fast-speculator}
There exists a speculator $\speculator$ that can be used in place of the target tool $\targettool$.
Given the same action $\action_i$, the speculator produces a speculative observation
$\hat{\observation}_i = \speculator(\action_i)$
in time $\Tspec < \Ttool$.
With probability $p>0$, this speculative observation induces the same next state as the target-tool observation
$\observation_i = \targettool(\action_i).$
That is, with probability $p$, we have
\[
\mathrm{Update}(\state_i,\seg_i,\action_i,\hat{\observation}_i)
=
\mathrm{Update}(\state_i,\seg_i,\action_i,\observation_i).
\]
\end{assumption}

\begin{assumption}[Verifiable equivalence]
\label{assump:verifiable-equivalence}
There exists a verifier $\verifier$ that can determine whether the speculative observation is equivalent to the target-tool observation for the purpose of the model's next state.
Formally, the verifier takes the current state, generated segment, action, speculative observation, and target observation, and returns
\[
\verifier(\state_i,\seg_i,\action_i,\hat{\observation}_i,\observation_i) \in \{0,1\}.
\]
When $\verifier(\state_i,\seg_i,\action_i,\hat{\observation}_i,\observation_i)=1$, the system is allowed to commit the speculative transition.
When the verifier returns $0$, the system has to discard the speculative branch.
\end{assumption}


\paragraph{Verification in information retrieval.}
In information-retrieval setups, verification can check whether the target tool and speculator provide the required information, e.g., the same factual knowledge or semantically equivalent evidence for the model's inquiry.
A restrictive verifier lowers the success rate $p$, while a looser verifier that ensures the speculative observation is sufficiently informative can preserve the trajectory to some extent and maintain task accuracy with higher $p$.

\subsection{Leveraging Speculation for Fast and Accurate Trajectories}

We parameterize the latency profile of the system using two dimensionless quantities.
First, we define the \emph{relative speculator latency}
$\alpha \triangleq {\expected{\Tspec}}/\expected{\Ttool}$.
Since the speculator is assumed to be faster than the target tool, we have $\alpha<1$.
Second, we define the \emph{decoding-to-tool latency ratio}
$\beta \triangleq \expected{\Tseg}/\expected{\Ttool}$.
This quantity measures how much time is spent on model-side segment generation relative to target-tool execution.



We first characterize the best theoretical latency gain achievable with the speculator $\speculator$.
To do so, we analyze an idealized oracle-verification setting that upper bounds the possible gain from speculation.

\begin{observationth}[Oracle speculative hop latency]
\label{thm:oracle-hop-latency}
Assume an idealized oracle-verification setting in which, at each hop, the system runs the speculator $\speculator$ and the target tool $\targettool$ in parallel.
As soon as the speculative observation is produced, an oracle verifier determines whether it would match the target-tool result, without waiting for the target tool to finish.
If speculation succeeds, which happens with probability $p$, the system proceeds using the speculative result; otherwise, it waits for and falls back to the target-tool result.
Then the expected latency of a single hop is
\[
\expected{\Thopspec}
=
\expected{\Tseg} + p\ \expected{\Tspec}
+
(1-p)\ \expected{\Ttool}.
\]
\end{observationth}

\begin{corollary}[Oracle relative latency]
\label{cor:oracle-relative-latency}
Under the setting of Observation~\ref{thm:oracle-hop-latency}, the relative latency of oracle speculation compared to sequential execution is
\[
\RelLatency^*
\triangleq
\frac{\expected{\Thopspec}}{\expected{\Thop}}
=
\frac{\expected{\Tseg}+p\expected{\Tspec}+(1-p)\expected{\Ttool}}{\expected{\Tseg}+\expected{\Ttool}}
=
1-\frac{p(1-\alpha)}{1+\beta}.
\]
\end{corollary}


Corollary~\ref{cor:oracle-relative-latency} shows that latency gain depends on three quantities:
the speculation success probability $p$, relative speculator latency $\alpha$, and decoding-to-tool latency ratio $\beta$.
Gains increase with more accurate or cheaper speculation, and decrease when model decoding dominates tool execution.

So far, we assumed an oracle verifier that immediately determines whether the target-tool output induces the same next state as the speculator.
In practice, verification requires observing the target-tool output and checking the speculative result.
\method{} addresses this issue through asynchronous verification while pushing speculation forward.

\subsection{\method{} with a Bounded Speculative Window}
\label{subsec:specpipe}

We describe an algorithm that maintains up to $k$ parallel threads, enabling speculation while verifying each speculative transition against the target tool.
Suppose the model is at state $\state_i$.
It generates the segment $\seg_i$ and action $\action_i$.
We create an original verifier thread \thread{1} that calls the target tool on $\action_i$.
In parallel, we invoke the speculator on the same action; using the speculative observation, the model transitions to a speculative next state and continues execution, creating a speculative continuation \thread{2}.
Then, \thread{2} generates the next segment and action, launches the target tool, and in parallel invokes the speculator again to create \thread{3}.
This process repeats: each thread calls the target tool for its current action, while the next thread is created by speculating that action's observation and continuing from the speculative state.
The system continues until it maintains $k$ active threads, denoted by \thread{1{:}k}.

When the target tool result from \thread{1} returns, the verifier $\verifier$ compares it with the speculative observation used to create \thread{2}.
If verification fails, then all \thread{(2{:}k)} are discarded, and the system resumes the thread \thread{1} for next hop at state $\state_{i+1}$.
If verification succeeds, then \thread{1} is discarded and \thread{(2{:}k)} remain valid continuations from the verified state.
The same procedure is then applied recursively: when the target-tool result from \thread{2} returns, it is compared against the speculative observation used to create \thread{3}.
If verified, \thread{2} is discarded and \thread{(3{:}k)} remain valid; otherwise, all threads are discarded and the system resumes \thread{2} from the verified state $\state_{i+2}$.
In general, the algorithm either verifies all speculative transitions and advances to $\state_{i+k}$, or encounters the first failed verification at some \thread{j}, discards all downstream threads \thread{(j+1{:}k)}, and resumes from the corresponding verified state $\state_{i+j}$.

Let $\Latency_k(N)$ denote the expected latency of \method{} with window size $k$ on a trajectory with $N$ remaining hops.
We characterize this latency in the following theorem.

\begin{theorem}[Expected runtime with a bounded speculative window]
\label{thm:bounded-window-runtime}
For a trajectory with $N$ remaining hops, suppose \method{} uses a speculative window of size $k$, and each speculative hop succeeds independently with probability $p$.
Then the expected wall-clock latency satisfies
\begin{align*}
    \Latency_k(N) = N \left(\expected{\Tseg} + \frac{\mu_k-1}{\mu_k}\expected{\Tspec} + \frac{1}{\mu_k}\expected{\Ttool} \right) + O(1),
\end{align*}
where $\mu_k = \frac{1-p^k}{1-p}$ is the expected number of hops advanced in one speculative round.
\end{theorem}

\begin{corollary}[Relative latency of bounded-window speculation]
\label{cor:bounded-window-relative-latency}
Under the setting of Theorem~\ref{thm:bounded-window-runtime}, the relative latency of \method{} with window size $k$ is
\[
\RelLatency_k
\triangleq
\frac{\Latency_k(N)}{N\ \expected{\Thop}}
=
\frac{\beta+\alpha+\frac{(1-\alpha)(1-p)}{1-p^k}}{1+\beta}
=
1-\frac{(1-\alpha)\left(1-\frac{1-p}{1-p^k}\right)}{1+\beta}
+
O\left(\frac{1}{N}\right).
\]
\end{corollary}

As shown in Corollary~\ref{cor:bounded-window-relative-latency}, \method{} approaches the ideal oracle performance as $k \to \infty$ when $N$ is sufficiently large.
In particular, $\lim_{k\to\infty} \RelLatency_k = \RelLatency^*.$
See Appendix~\ref{app:proof-runtime} for the proof.

\begin{algorithm}[t]
\caption{\method{}: Continuous Speculative Execution with $k$ Active Threads}
\label{alg:abstract-continuous-speculation}
\footnotesize
\begin{algorithmic}[1]
\Require query $q$, model $\generatormodel$, target tool $\targettool$, speculator $\speculator$, verifier $\verifier$, window size $k$
\Ensure Final answer $f$

\State $\textsc{T}_1 \gets \mathrm{Init}(M, q)$
\State $\mathcal{W} \gets \{\textsc{T}_1\}$ \Comment{Initialize with the first thread}

\While{true}
    \If{$|\mathcal{W}| < k$} \Comment{Extend the speculative window}
        \State Extend the last thread in $\mathcal{W}$ by one hop using model $\generatormodel$ and the speculator $\speculator$
        \State Launch the corresponding target-tool call $\targettool$ asynchronously
    \EndIf

    \Statex \Comment{Verify speculation with target-tool observations}
    \If{the earliest target-tool call in $\mathcal{W}$ has returned}
        \State Verify its result against the corresponding speculative observation using $\verifier$

        \If{verification succeeds}
            \State Commit the earliest speculative hop and shift $\mathcal{W}$ forward
        \Else
            \State Roll back to the verified thread with target-tool called
            \State $\mathcal{W} \gets \{\textsc{T}_{\mathrm{verified}}\}$ \Comment{Keep the verified thread}
        \EndIf
    \EndIf

    \If{the verified trajectory reaches a final answer}
        \State \Return $f$
    \EndIf
\EndWhile
\end{algorithmic}
\end{algorithm}

\subsection{\method{} with Continuous Speculation but Bounded Active Threads}
\label{sec:cont-spec}

Motivated by Corollary~\ref{cor:bounded-window-relative-latency}, we introduce a continuous version of \method{} that avoids the stop-and-wait bottleneck of a fixed speculative window.
We first describe a variant that maintains $k$ active threads throughout execution, and then discuss how to choose $k$ to approximate the $k\to\infty$ behavior.
In the bounded-window version, the system creates up to $k$ threads and then waits for verification before deciding which speculative trajectory to retain.
When speculation is consistently successful, this can still create a bottleneck: once the window is exhausted, the system may wait for the target-tool output of the last thread before creating further speculative continuations.

The continuous version avoids this by keeping $k$ active threads at all times.
Whenever the earliest unresolved speculation is verified, the corresponding verifier thread is discarded, the window shifts forward, and the system immediately extends the frontier by speculating from the last active thread.
If verification fails, all downstream speculative threads are discarded and execution resumes from the verified target-tool state.
Thus, as long as speculation remains valid, the system keeps speculating ahead rather than stopping at the end of a finite window.
Algorithm~\ref{alg:abstract-continuous-speculation} gives an abstract implementation, with full implementation-level pseudocode in Appendix~\ref{alg:continuous-window}.

\paragraph{Selecting $k$ for \method{} with continuous speculation.}

The continuous version of \method{} does not need an unbounded number of active threads to emulate the $k\to\infty$ behavior.
The key observation is that a new speculative thread is only useful if it can be created before the earliest unresolved target-tool call returns.
Otherwise, the system would already have verified the earliest speculation and shifted the window forward.
Threads opened during that wait are blocked on $\targettool$, so concurrent calls to $\generatormodel$ stay about 2 in our runs rather than growing with $k$ (Appendix~\ref{app:live}), and per-call times increase only slightly (Table~\ref{tab:live-contention}).
To provide theoretical insight into how many active threads are required, we let $\nu$ denote the maximum coefficient of variation of the system, defined such that for any latency variable $X \in \{\Tseg, \Tspec, \Ttool\}$, its standard deviation is bounded by its expectation: $\mathrm{std}(X) \le \nu \expected{X}.$

\begin{theorem}[Probability of \method{} with $k$ active threads starves]
\label{thm:starvation-prob}
Under a CLT approximation, for \method{} with $k$ active speculative threads, the probability that the pipeline starves before the target tool resolves is bounded using the standard normal CDF $\Phi$:
\[
P_{\text{starve}}(k) \le \Phi \left( \frac{(1 + \beta) - k(\alpha + \beta)}{\nu \sqrt{k\alpha^2 + (k-1)\beta^2 + 1}} \right)
\]
\end{theorem}



According to Theorem~\ref{thm:starvation-prob} (see Appendix~\ref{app:proof-starve} for proof and details), choosing
$k^\star=\left\lceil{(1+\beta)}/{(\alpha+\beta)}\right\rceil$
ensures $P_\text{starve}\leq 0.5$.
To illustrate what a reliable choice of $k$ looks like in practice, we consider two representative settings motivated by our empirical measurements, aiming for a starvation probability below $5\%$ under the volatility bound $\nu\leq0.4$.
Using Theorem~\ref{thm:starvation-prob}, when $(\alpha,\beta)=(0.2,0.15)$, choosing $k^\star_{0.05}=6$ satisfies this goal; when $(\alpha,\beta)=(0.3,0.75)$, choosing $k^\star_{0.05}=3$ suffices.
Thus, near-optimal latency can be achieved with a small number of threads in realistic stochastic settings.




\section{Experiments}
\label{sec:experiments}

In this section, we empirically evaluate \method{}.
We first describe the experimental setup in Section~\ref{sec:exp-setup}.
We then present our main results in Section~\ref{sec:results}, showing that \method{} substantially reduces wall-clock latency.
Additional analyses and ablations can be found in Appendix~\ref{app:add-exp}).


\input{tables/main_rellat}

\subsection{Experimental Setup}
\label{sec:exp-setup}

\paragraph{\textbf{Datasets.}} We evaluate \method{} on three multi-hop question-answering datasets: 2WikiMultihopQA~\cite{ho2020constructing}, MuSiQue~\cite{trivedi2022musique}, and DeepResearch-9K~\cite{wu2026deepresearch}.
For 2WikiMultihopQA and MuSiQue, the per-example hop budget is derived from their dataset attributes (i.e., number of distinct Wikipedia titles appearing in \texttt{supporting\_facts} for 2WikiMultihopQA, and length of \texttt{question\_decomposition} for MuSiQue), ranging from 2 to 4 hops.
We allow our model the budget of 2 extra hops for each sample. For DeepResearch-9K we evaluate samples with difficulty 2 (difficulty ranges from 1 to 3), as they provide challenging tasks that need multiple hops of retrieval for answering, and are not too complex for models in our parameter range, and we allow fix budget of 10 hops for samples.


\paragraph{\textbf{Generator Model ($\generatormodel$).}}
Our primary generator is CoRAG-Llama3.1-8B-MultihopQA~\cite{wang2025corag}, which is trained to generate intermediate \subq{}s hop-by-hop, making it a natural fit for continuous speculative execution.
Appendix~\ref{app:gpt_baseline} also reports additional experiments with GPT-5~\cite{singh2025openai} as $\generatormodel$.

\paragraph{\textbf{Target Tool ($\targettool$).}}
We evaluate two distinct retrieval backends as our target tools. {(1) E5 Retrieval:} A local HTTP service querying a KILT/Wikipedia passage corpus, returning the top 5 nearest-neighbor passages for each \subq{}
{(2) Web Search:} A DuckDuckGo-based web retrieval tool returning 5 search results per query, with a 3-second page-fetch timeout. 
    

\paragraph{\textbf{Speculator ($\speculator$) and verifer ($\verifier$)}}
We mainly use LLMs as speculator, including Llama 3.1 8B~\cite{grattafiori2024llama}, Qwen 3 8B~\cite{Yangetal2025}, GPT-4o mini, and GPT-4o~\cite{hurst2024gpt}.
These models span different speed--accuracy trade-offs, inducing distinct $(p,\alpha)$ values.
Since \method{} only requires a fast substitute for the target tool $\targettool$, we also evaluate a cached retrieval backend as the speculator in Section~\ref{sec:results}.
We use a deterministic rule-based verifier that compares the target-tool \suba{} with the speculative \suba{} for each hop and decides whether to accept the speculation.
The verifier relies on normalization, exact match, and token-set Jaccard similarity.
Appendix~\ref{app:exp-setup} provides implementation details, and Appendix~\ref{app:verifier-ablation} reports a human evaluation of its reliability.



\input{tables/em_f1_latency}

\paragraph{\textbf{Standard multi-hop execution.}}
Given an input question, $\generatormodel$ generates a \subq{} at each hop.
The target tool $\targettool$ retrieves top documents containing the answer to the \subq{}, and $\generatormodel$ uses them to produce a \suba{}.
The resulting (\subq{}, \suba{}) pair is added to the context for generating the next \subq{}.
After the hop budget is reached, $\generatormodel$ uses the original question and all $N$ (\subq{}, \suba{}) pairs to generate the final answer.



Appendix~\ref{app:exp-setup} contains more details on the experimental setup, including generation hyperparameters, specific target tool configurations, early stopping criteria, and how latency is measured.

In our experiments, a single thread replays the events of every speculative thread and closely approximates the latency that parallel execution would have had.
Parallel execution can still add overhead from contention on calls, which we measure in Appendix~\ref{app:live}.

\subsection{Results}
\label{sec:results}



\input{figures/k_rellat}

We run continuous \method{} with $k \rightarrow \infty$ setting, meaning that we do not impose a limit on the number of active threads.
The algorithm dynamically creates new threads until the first unresolved thread is verified, allowing speculative continuations to yield additional latency reduction.

\paragraph{\textbf{\method{} achieves significant latency reduction across settings.}}

We use hat notation (e.g., $\hat{p}$, $\hat{\alpha}$, $\hat{\beta}$) to denote empirical values measured in our experiments, distinguishing them from their theoretical counterparts.
As shown in Table~\ref{tab:spec-metrics-three-datasets}, \method{} achieves latency gains that closely match the optimal theoretical limits ($\RelLatency^*$) across three datasets, two target tools $\targettool$, and diverse speculators $\speculator$.
In the Web search setting, network-bound fetching creates a large bottleneck, making local speculative generation much faster and yielding low relative speculator latency ($\hat{\alpha}$).
GPT-based speculators incur API overhead and thus have higher $\hat{\alpha}$ than local Llama and Qwen models, but they achieve higher speculation success rates $\hat{p}$.
We omit GPT speculators for the E5 target tool as their API latency is not consistently lower than local E5 retrieval latency ($\hat{\alpha}>1$), violating faster speculator assumption.

Overall, \method{}'s empirical relative latency ($\RelLatency$) tightly approaches the optimal bounds.
For example, using GPT-4o as the Web-search speculator on 2WikiMultihopQA achieves $\RelLatency=0.60$, corresponding to a $40\%$ reduction in total execution time, close to its optimal limit of $0.50$.

\paragraph{\textbf{Comparison to one-step speculation.}}
Speculative Actions~\cite{ye2025speculative} launches the next tool call from a speculative observation and then waits for the current target call to resolve.
Table~\ref{tab:spec-actions} compares that schedule with \method{} on the same web-search traces.
At $k=2$, \method{} is lower by $0.04$, $0.03$, and $0.03$ relative latency on the three datasets.
The gap is larger at $k \rightarrow \infty$ ($0.60$ vs.\ $0.78$ on 2WikiMultihopQA).
Speculative Actions uses fewer $\generatormodel$ calls because it does not keep a window open (Table~\ref{tab:spec-actions-calls}).

\input{tables/spec_actions}


\paragraph{\textbf{\method{} fully preserves trajectory and answer quality.}}

As detailed in Tables~\ref{tab:spec-metrics-absolute-relative}, \method{} tightly preserves the task accuracy and final answer quality of the standard sequential execution.
Across all datasets and speculator configurations, deviations in EM and F1 scores (Exact Match and token F1 on SQuAD-normalized answers) are marginal.
By employing a strictly defined verifier ($\verifier$), \method{} ensures that no unauthorized deviations occur in the multi-hop trajectory. This verification step keeps the system consistent.
On a blind check of 500 decisions across datasets, target tools, and speculators, verifier precision is $0.996$ and recall is $0.91$ (Appendix~\ref{app:verifier-ablation}).
Conversely, the ``Full Speculation'' approach, i.e., for all hops only the speculator is called, demonstrates the danger of unverified generation.
While this method is significantly faster (particularly when bypassing an expensive $\targettool$ like Web Search) it heavily degrades answer quality.
As shown in Table~\ref{tab:spec-metrics-absolute-relative}, Full Speculation causes EM to plummet from 68.7 to 38.7 on 2WikiMultihopQA and nearly halves it from 10.0 to 5.7 on MuSiQue.
\method{} effectively eliminates this accuracy trade-off, securing substantial speedups while maintaining reliability.


\paragraph{\textbf{\method{} offers a latency-compute cost trade-off via $k$.}}

Theorem~\ref{thm:starvation-prob} provides insights on how increasing $k$ results in smaller chance of starvation.
However, maintaining a large number of active threads inherently increases the total number of parallel executions.
By restricting \method{} to smaller active threads, it provides a flexible trade-off between maximizing latency reduction and controlling computational overhead.
Figure~\ref{fig:k_rellat} illustrates this dynamic for the continuous \method{} framework, utilizing GPT-4o as the speculator ($\speculator$) and Web Search as the target tool ($\targettool$).
The plots track both the relative latency and the total number of model/tool calls across different values of $k$. 

For instance, on DeepResearch-9K, restricting the window to $k=3$ achieves a highly competitive $\RelLatency$ of $0.75$, close to the $k \rightarrow \infty$ limit of $0.71$.
At this reduced capacity, the system incurs a moderate compute overhead: the average number of $\targettool$ calls increases from $5.48$ in standard sequential execution to $10.69$, while $\generatormodel$ calls increase from $5.48$ to $17.83$.
Pricing those DeepResearch-9K call counts at the API rates in Table~\ref{tab:api-cost} gives \$0.06 per query for sequential execution and \$0.16 at $k=3$.
Finally, we note that while our theoretical analysis focuses on avoiding starvation and derives bounds based on $(\hat{\alpha},\hat{\beta})$, the shape of this trade-off curve is also strongly influenced by the speculator's empirical success probability $\hat{p}$ and behavior; when the speculator is weak, larger values of $k$ are less useful.
Evaluation of this trade-off across additional configurations, including GPT-5 as $\generatormodel$, is provided in Figure~\ref{fig:k_rellat_appendix_all}.

\input{figures/e5_cache}

\paragraph{\textbf{\method{} is not limited to LLM speculators.}}
In retrieval-augmented systems, a small optimized cache can answer queries much faster than full retrieval or web search, but a naive cache-first strategy is unreliable because incomplete or misleading cache results can compound errors.
\method{} instead uses the cache as a speculator, using its low latency while verifying outputs against the target tool $\targettool$.

We implement a fast E5-cache backend containing only 5\%, 10\%, or 25\% of the Wikipedia index, selecting ``heavy-hitter'' passages by Wikipedia in-degree as a proxy for centrality.
As shown in Figure~\ref{fig:e5_cache_2wiki}, increasing the cache fraction from 0.05 to 0.25 steadily improves $\hat{p}$, exceeding 60\% for E5 retrieval and 50\% for Web Search.
Since the local cache is much faster than web search ($\hat{\alpha}\approx0.05$), this yields substantial speedups; at 0.25 cache fraction, \method{} reduces Web Search relative latency to $\RelLatency=0.64$,
demonstrating that simple, high-speed data subsystems can seamlessly accelerate complex reasoning agents when paired with continuous verification.

\section{Limitations}

\method{} requires a fast, approximate speculator ($\speculator$) and a reliable verifier ($\verifier$), as formalized in Assumptions~\ref{assump:fast-speculator} and \ref{assump:verifiable-equivalence}. In domains where these tools are unavailable, or where verifying an answer is as expensive as executing the full target tool ($\targettool$), the latency reduction from continuous speculation cannot be achieved. Furthermore, while \method{} achieves significant latency reduction in many settings, it generally introduces a noticeable increase in the number of $\generatormodel$, $\speculator$, and $\targettool$ calls. Nevertheless, \method{} remains a highly practical solution in scenarios where minimizing wall-clock latency is critical and a higher computational overhead is a tolerable trade-off.

\section{Conclusion}

We accelerate multi-hop information-intensive trajectories by proposing \method{}, which leverages a speculator to continuously advance speculative execution. We provide a theoretical framework to study the latency of our algorithm and compare it with the optimal achievable gain. Empirically, we show that \method{} closely follows the theory and reduces latency while preserving task accuracy.

\section*{Acknowledgement}
This project was supported in part by a grant from an NSF CAREER AWARD 1942230, the ONR PECASE grant N00014-25-1-2378, ARO’s Early Career Program Award 310902-00001, Army Grant No. W911NF2120076, the NSF award CCF2212458, NSF Award No. 2229885 (NSF Institute for Trustworthy AI in Law and Society, TRAILS), a MURI grant 14262683, DARPA AIQ grant HR00112590066 and an award from meta 314593-00001.

\bibliographystyle{plainnat}
\bibliography{ref}

\newpage
\appendix

\input{appendix}


\end{document}

%% file: figures/teaser.tex
\renewcommand{\thread}[1]{\textsc{T}_{#1}}
\begin{figure*}[t]
    \centering
    \includegraphics[width=\textwidth]{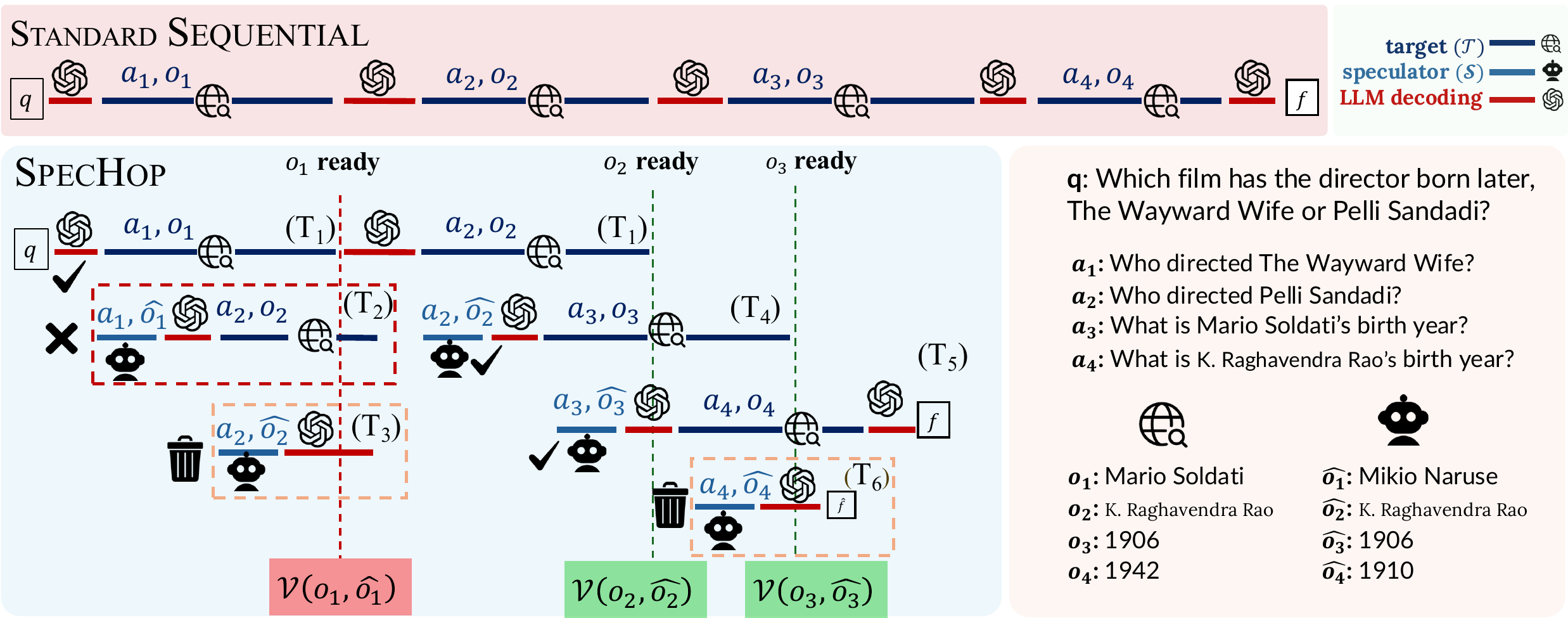}
    \vspace{-10pt}
\caption{
Overview of \method{} with \textbf{continuous speculative execution}, maintaining $k=3$ active threads on a query requiring $4$ external calls.
Initially, $\thread{1}$ calls the target tool for $(\action_1,\observation_1)$.
In parallel, $\thread{2}$ speculates the first observation as $\hat{\observation}_1$ and continues to the next hop, where it calls the target tool for $(\action_2,\observation_2)$; $\thread{3}$ is created similarly.
When $\observation_1$ returns, the verifier $\verifier$ compares it with $\hat{\observation}_1$.
Since verification fails, $\thread{2}$ and its downstream speculative thread $\thread{3}$ are discarded, and the method resumes from the last verified thread.
The system then creates new speculative threads $\thread{4}$ and $\thread{5}$ while $\thread{1}$ waits for the target output of $(\action_2,\observation_2)$.
When $\observation_2$ returns, verification against $\hat{\observation}_2$ from $\thread{4}$ succeeds, so the system commits to $\thread{4}$ and continues speculation with $\thread{6}$.
Subsequent verifications also succeed: the system commits to $\thread{5}$ after $\observation_3$ returns, and once $\observation_4$ returns, $\thread{5}$ produces the final answer $f$.
By keeping the pipeline active instead of waiting sequentially at each hop, \method{} reduces wall-clock latency while preserving the original verified trajectory.
}
    \label{fig:specpipe-overview}
    \vspace{-10pt}
\end{figure*}

\renewcommand{\thread}[1]{$\textsc{T}_{#1}$}

%% file: figures/state_machine.tex
\begin{center}
    \begin{tikzpicture}[
        state/.style={circle, draw, thick, minimum size=0.65cm, inner sep=1pt},
        finalstate/.style={circle, draw, thick, double, minimum size=0.65cm, inner sep=1pt},
        arrow/.style={->, thick},
        every node/.style={font=\small}
    ]
        \node (q) {$q$};
        \node[state, right=0.3cm of q] (s0) {$\state_0$};

        \node[right=0.4cm of s0] (dots1) {$\cdots$};

        \node[state, right=0.4cm of dots1] (si) {$\state_i$};
        \node[state, right=2.4cm of si] (sip1) {$\state_{i+1}$};

        \node[right=0.4cm of sip1] (dots2) {$\cdots$};

        \node[finalstate, right=0.4cm of dots2] (sn) {$\state_N$};
        \node[right=0.3cm of sn] (f) {$f$};

        \draw[arrow] (q) -- (s0);

        \draw[arrow] (s0) -- (dots1);
        \draw[arrow] (dots1) -- (si);

        \draw[arrow] (si) -- (sip1)
            node[midway, above, font=\footnotesize] {$\left(\seg_i,\left(\action_i,\observation_i\right)\right)$}
            node[midway, below, font=\footnotesize] {hop};

        \draw[arrow] (sip1) -- (dots2);
        \draw[arrow] (dots2) -- (sn);

        \draw[arrow] (sn) -- (f);
    \end{tikzpicture}
\end{center}

%% file: tables/main_rellat.tex

\definecolor{metricbg}{HTML}{F4F7F9} 
\definecolor{bestval}{HTML}{0A3B5C}  
\newcommand{\best}[1]{\textcolor{bestval}{\textbf{#1}}} 

\begin{table*}[t]
  \centering
  \renewcommand{\arraystretch}{1.5}
  \setlength{\tabcolsep}{3pt} 
  \caption{Latency reduction achieved by \method{}. We report the empirical speculation success probability ($\hat{p}$), relative speculator latency ($\hat{\alpha}$), and the decoding-to-tool latency ratio ($\hat{\beta}$). \method{} consistently achieves empirical relative latencies ($\RelLatency$) that closely approach the theoretical optimal ($\hat{\RelLatency}^*$) across three datasets, two target tools ($\targettool$), and various speculative models ($\speculator$).}
  \vspace{2pt}
  \resizebox{\textwidth}{!}{%
  \begin{tabular}{@{}c l ccc >{\columncolor{metricbg}}c >{\columncolor{metricbg}}c ccc >{\columncolor{metricbg}}c >{\columncolor{metricbg}}c ccc >{\columncolor{metricbg}}c >{\columncolor{metricbg}}c@{}}
    \toprule
    & & \multicolumn{5}{c}{\textbf{2WikiMultihopQA}~\cite{ho2020constructing}} 
    & \multicolumn{5}{c}{\textbf{MuSiQue}~\cite{trivedi2022musique}} 
    & \multicolumn{5}{c}{\textbf{DeepResearch-9K}~\cite{wu2026deepresearch}} \\
    
    \cmidrule(lr){3-7} \cmidrule(lr){8-12} \cmidrule(lr){13-17}
    
    Target Tool $(\targettool)$ & Speculator ($\mathcal{S}$)
      & $\hat{p}$ & $\hat{\alpha}$ & $\hat{\beta}$ & $\hat{\RelLatency}^*$ $\downarrow$ & $\RelLatency$ $\ \downarrow$
      & $\hat{p}$ &  $\hat{\alpha}$ & $\hat{\beta}$ & $\hat{\RelLatency}^*$ $\downarrow$ & $\RelLatency$ $\ \downarrow$
      & $\hat{p}$ &  $\hat{\alpha}$ & $\hat{\beta}$ & $\hat{\RelLatency}^*$ $\downarrow$ & $\RelLatency$ $\ \downarrow$ \\
    \midrule \midrule
    
    \multirow{2}{*}{{E5 Retrieval}}
      & Llama 3.1 8B
        & 0.27 & 0.30 & 0.74 & \best{0.89} & \best{0.91}
        & 0.42 & 0.30 & 0.64 & \best{0.82} & \best{0.86}
        & 0.39 & 0.29 & 0.82 & \best{0.85} & \best{0.88} \\
      & Qwen 3 8B
        & 0.15 & 0.35 & 0.74 & 0.95 & 0.95
        & 0.34 & 0.29 & 0.64 & 0.85 & 0.89
        & 0.25 & 0.30 & 0.81 & 0.90 & 0.92 \\
    \midrule
    
    \multirow{4}{*}{{Web Search}}
      & Llama 3.1 8B
        & 0.28 & 0.03 & 0.10 & 0.75 & 0.78
        & 0.41 & 0.05 & 0.11 & 0.65 & 0.72
        & 0.35 & 0.05 & 0.13 & 0.70 & 0.76 \\
      & Qwen 3 8B
        & 0.18 & 0.05 & 0.10 & 0.85 & 0.88
        & 0.33 & 0.05 & 0.11 & 0.72 & 0.78
        & 0.31 & 0.05 & 0.13 & 0.74 & 0.78 \\
      & GPT-4o mini
        & 0.41 & 0.16 & 0.10 & 0.68 & 0.76
        & 0.49 & 0.20 & 0.11 & 0.65 & 0.72
        & 0.44 & 0.22 & 0.13 & 0.69 & 0.74 \\
      & GPT-4o
        & 0.68 & 0.19 & 0.10 & \best{0.50} & \best{0.60}
        & 0.50 & 0.19 & 0.11 & \best{0.63} & \best{0.69}
        & 0.45 & 0.18 & 0.13 & \best{0.68} & \best{0.71} \\
    \bottomrule
  \end{tabular}%
  }
    
    \label{tab:spec-metrics-three-datasets}
\end{table*}

%% file: tables/em_f1_latency.tex

\definecolor{metricbg}{HTML}{F4F7F9} 

\begin{table*}[t]
  \centering 
  \renewcommand{\arraystretch}{1.0}
  \setlength{\tabcolsep}{3pt} 
  \footnotesize
  \caption{End-to-end task performance (EM/F1) and absolute wall-clock latency (seconds). \method{} preserves the accuracy of the standard trajectory (which uses $\generatormodel$ and $\targettool$ without $\speculator$), while substantially reducing execution time. Here, $\speculator$ is Llama 3.1 8B. Full Speculation uses only the speculator for all hops without target-tool verification, leading to severe accuracy degradation.}
  \vspace{2pt}
  \begin{tabular}{@{} c l cc c cc c cc c @{}}
    \toprule
    & & \multicolumn{3}{c}{\textbf{2WikiMultihopQA}~\cite{ho2020constructing}} & \multicolumn{3}{c}{\textbf{MuSiQue}~\cite{trivedi2022musique}} & \multicolumn{3}{c}{\textbf{DeepResearch-9K}~\cite{wu2026deepresearch}} \\
    \cmidrule(lr){3-5} \cmidrule(lr){6-8} \cmidrule(lr){9-11}
    Target Tool $(\targettool)$& Method 
      & EM & F1 & Latency (s) 
      & EM & F1 & Latency (s) 
      & EM & F1 & Latency (s) \\
    \midrule \midrule
    \multirow{2}{*}{{Web Search}}
      & Standard
        & 68.7 & 73.8 & 21.13
        & 10.0 & 25.1 & 15.59
        & 32.0 & 35.6 & 20.17 \\
      & \textbf{\method{}} 
        & 69.3 & 73.8 & 16.10
        & 10.7 & 25.3 & 11.17
        & 32.0 & 35.6 & 15.07 \\
    \midrule
    \multirow{2}{*}{{E5 Retrieval}}
      & Standard
        & 72.3 & 76.9 & 3.81
        & 23.7 & 36.5 & 3.89
        & 22.0 & 30.1 & 5.49 \\
      & \textbf{\method{}} 
        & 72.3 & 76.9 & 3.47
        & 23.7 & 36.6 & 3.37
        & 22.3 & 30.3 & 4.83 \\
    \midrule
    \multirow{1}{*}{--}
      & Full Speculation
        & 38.7 & 42.1 & 2.46
        & 5.7 & 15.4 & 2.35
        & 12.3 & 16.9 & 3.66 \\
    \bottomrule
  \end{tabular}%
    \label{tab:spec-metrics-absolute-relative}
\end{table*}

%% file: figures/k_rellat.tex
\begin{figure}[t]
    \centering
    \includegraphics[width=0.9\linewidth]{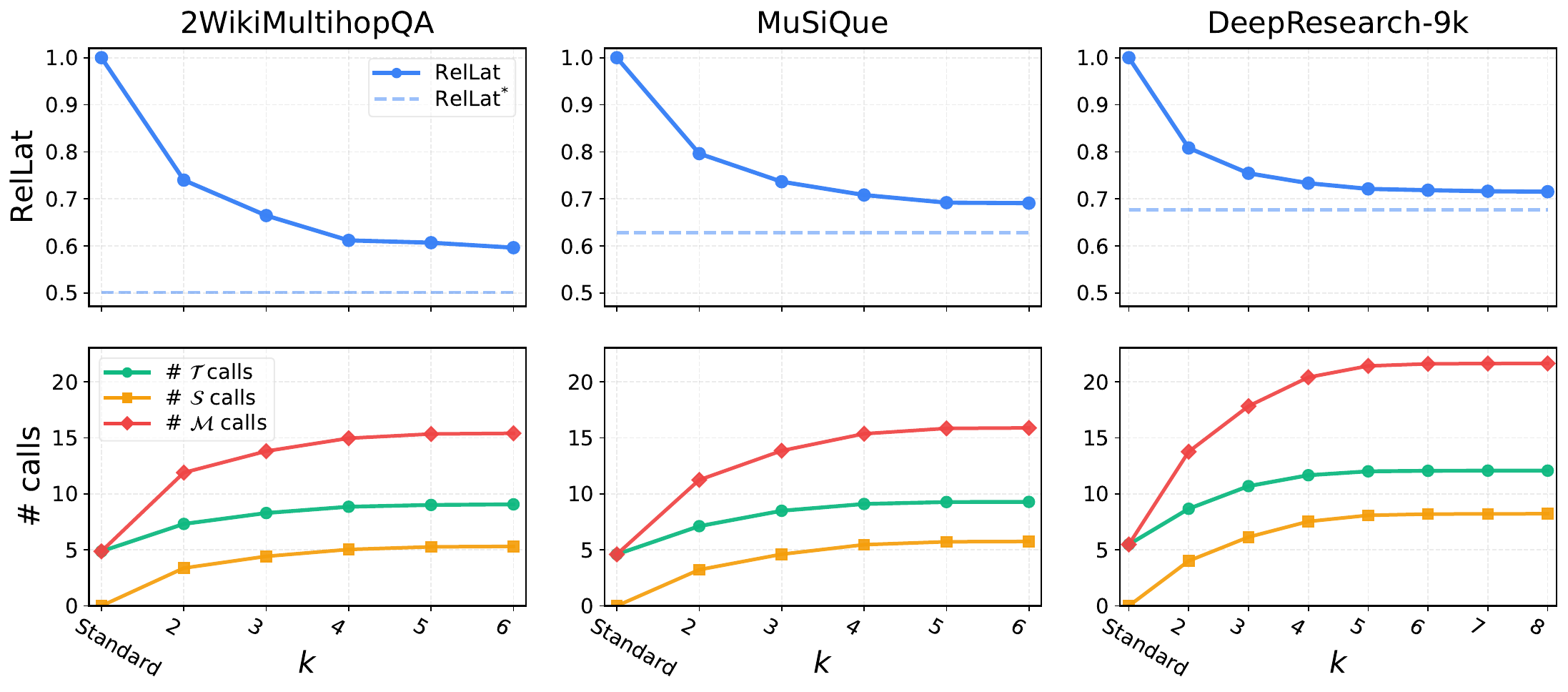}
    \vspace{-10pt}
    \caption{The effect of active thread limit ($k$) on relative latency ($\RelLatency$) and computational cost (average number of calls to the target tool $\targettool$, speculator $\speculator$, and generator model $\generatormodel$). The evaluation setting uses GPT-4o as the speculator ($\speculator$) and Web Search as the target tool ($\targettool$). The dashed line is the theoretical optimal relative latency ($\RelLatency^*$). The ``Standard'' setting refers to not using \method{}. 
    Additional trade-off results for other configurations are provided in Appendix (Figure~\ref{fig:k_rellat_appendix_all}).}
    \label{fig:k_rellat}
\end{figure}

%% file: tables/spec_actions.tex
\begin{table}[t]
  \centering
  \small
  \renewcommand{\arraystretch}{1.15}
  \setlength{\tabcolsep}{5pt}
  \caption{Relative latency of Ye et al.~\cite{ye2025speculative} and \method{} on the same web-search traces (CoRAG as $\generatormodel$, GPT-4o as $\speculator$, 150 examples per dataset). Ye et al. speculate one hop ahead and then wait. Call counts are in Table~\ref{tab:spec-actions-calls}.}
  \vspace{2pt}
  \begin{tabular}{@{}l ccc@{}}
    \toprule
    Method & 2Wiki & MuSiQue & DeepResearch-9K \\
    \midrule
    Standard & 1.00 & 1.00 & 1.00 \\
    Ye et al.~\cite{ye2025speculative} & 0.78 & 0.83 & 0.84 \\
    \method{}, $k=2$ & 0.74 & 0.80 & 0.81 \\
    \method{}, $k=3$ & 0.66 & 0.74 & 0.75 \\
    \method{}, $k \rightarrow \infty$ & 0.60 & 0.69 & 0.71 \\
    \bottomrule
  \end{tabular}
  \label{tab:spec-actions}
\end{table}

%% file: figures/e5_cache.tex
\begin{figure}[t]
  \centering
  \includegraphics[width=0.9\linewidth]{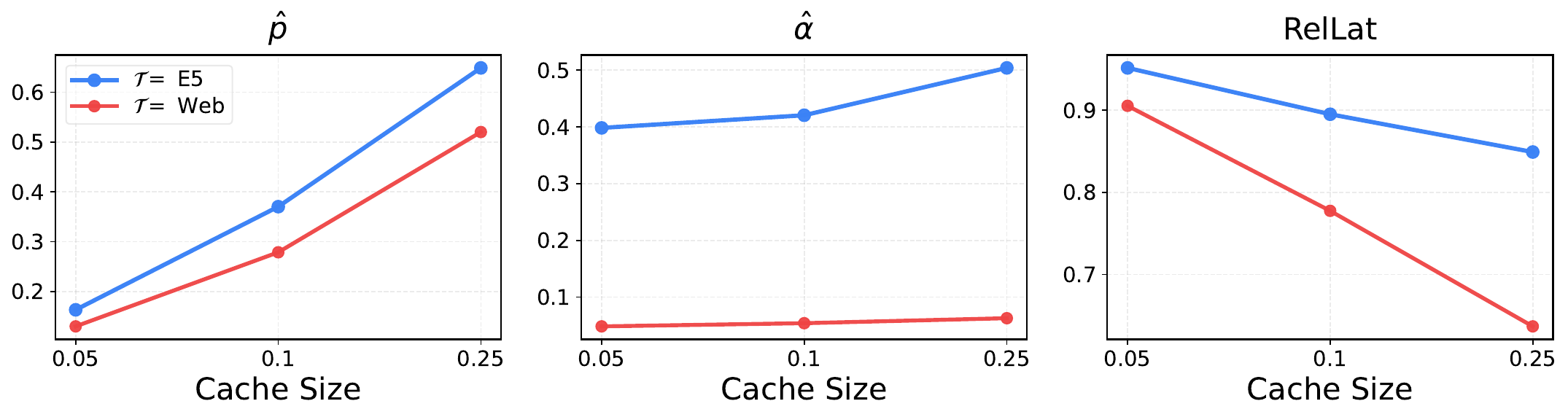}
  \vspace{-5pt}
  \caption{Performance of \method{} using a fast E5-cache as the speculator ($\speculator$) on the 2WikiMultihopQA dataset. The cache size varies from 5\% to 25\% of the full index.
  The plots show the empirical success probability ($\hat{p}$), relative speculator latency ($\hat{\alpha}$), and resulting relative latency ($\RelLatency$) when accelerating both the full E5 retriever and Web Search target tools ($\targettool$).}
  \label{fig:e5_cache_2wiki}
  \vspace{-10pt}
\end{figure}

%% file: appendix.tex
\section{Full Implementation of the Algorithm}
\label{app:pseudo-code}

\begin{algorithm}[H]
\caption{\method{}: Continuous Speculative Execution with $k$ Active Threads}
\label{alg:continuous-window}
\small
\begin{algorithmic}[1]
\Require Query $q$, model $M$, target tool $T$, speculator $S$, verifier $V$, window size $k$
\Ensure Final answer $f$

\State $\state \gets \mathrm{Init}(M,q)$

\While{true}

    \State $(\seg,\action) \gets M(\state)$
    \If{$\seg$ contains a final answer}
        \State \Return $\mathrm{ExtractAnswer}(\seg)$
    \EndIf

    \State Launch $T(\action)$ asynchronously
    \State $\mathrm{verifier\_thread} \gets (\state,\seg,(\action,T(\action)))$
    \State $\mathcal{W} \gets [\mathrm{verifier\_thread}]$

    \While{true}

        \If{$|\mathcal{W}| < k$}
            \State $(\state',\seg',(\action',\cdot)) \gets \mathrm{Back}(\mathcal{W})$

            \If{$\seg'$ does contain a final answer}
                \State $\hat{\observation}' \gets S(\action')$
                \State $\hat{\state}' \gets \mathrm{Update}(\state',\seg',\action',\hat{\observation}')$
                \State $(\seg_{\mathrm{spec}},\action_{\mathrm{spec}}) \gets M(\hat{\state}')$
                \State Launch $T(\action_{\mathrm{spec}})$ asynchronously
                \State $\mathrm{spec\_thread} \gets (\hat{\state}',\seg_{\mathrm{spec}},(\action_{\mathrm{spec}},T(\action_{\mathrm{spec}})))$
                \State $\mathcal{W}.\mathrm{push\_back}(\mathrm{spec\_thread})$
            \EndIf
        \EndIf

        \If{$\mathrm{verifier\_thread}$ is done}
            \State $(\state_0,\seg_0,(\action_0,\observation_0)) \gets \mathrm{verifier\_thread}$
            \State $\mathcal{W}.\mathrm{pop\_front}()$
            \State $\mathrm{spec\_thread} \gets \mathrm{Front}(\mathcal{W})$

            \State $\left(\left(\state_0,\seg_0,(\action_0,\hat{\observation}_0)\right),\seg_{\mathrm{spec}},(\action_{\mathrm{spec}},\cdot)\right) \gets \mathrm{spec\_thread}$

            \If{$V(\observation_0,\hat{\observation}_0)=\mathrm{true}$}
                \State $\mathrm{verifier\_thread} \gets \mathrm{spec\_thread}$
            \Else
                \State $\state \gets \mathrm{Update}(\state_0,\seg_0,\action_0,\observation_0)$
                \State $\mathcal{W} \gets \emptyset$
                \State \textbf{break}
            \EndIf
        \EndIf

    \EndWhile

\EndWhile
\end{algorithmic}
\end{algorithm}

\section{Proofs}
\label{app:proofs}

\subsection{Proof of Theorem~\ref{thm:bounded-window-runtime}}
\label{app:proof-runtime}

Let $\Latency_k(N)$ denote the expected wall-clock latency of \method{} with a bounded speculative window of size $k$ on a trajectory with $N$ remaining hops.
We assume the latencies of individual segment generations, speculative tool calls, and target tool calls are independent observations of the random variables $\Tseg$, $\Tspec$, and $\Ttool$, respectively.

For $1 \leq j < k$, the system advances by $j$ hops when the first $j-1$ speculative hops are verified and the $j$-th speculative hop fails. This event has probability $p^{j-1}(1-p)$.
For $j=k$, the system advances by $k$ hops when the first $k-1$ speculative hops are all verified, which has probability $p^{k-1}$.
We can compact these probabilities by defining the weight sequence:
\[
w_j =
\begin{cases}
p^{j-1}(1-p), & 1 \leq j < k, \\
p^{k-1}, & j=k.
\end{cases}
\]
These weights form a valid probability distribution for the number of hops advanced in a single round, summing to one:
\[
\sum_{j=1}^k w_j
=
\sum_{j=1}^{k-1} p^{j-1}(1-p)+p^{k-1}
=
1.
\]

If the system advances by $j$ hops, it must sequentially wait for the generation of $j$ segments, $j-1$ speculative tool calls,
and exactly one target-tool call to resolve the final state.
Because expectation is a linear operator, the expected wall-clock cost of this round is:
\[
\expected{C_j} = j\expected{\Tseg} + (j-1)\expected{\Tspec} + \expected{\Ttool}.
\]

By the Law of Total Expectation, we can formulate the recurrence relation for the remaining trajectory as:
\[
\Latency_k(N)
=
\sum_{j=1}^k w_j\big(\expected{C_j} + \Latency_k(n-j)\big).
\]

We now solve this recurrence asymptotically.
Assume
\[
\Latency_k(N)=\lambda_k\ N+O(1),
\]
where $\lambda_k$ is the expected latency per completed hop.
Substituting into the recurrence gives:
\begin{align*}
\lambda_k\ N+O(1)
&=
\sum_{j=1}^k w_j\left(\expected{C_j} + \lambda_k(N-j) + O(1)\right) \\
&=
\sum_{j=1}^k w_j \expected{C_j}
+
\lambda_k N\sum_{j=1}^k w_j
-
\lambda_k\sum_{j=1}^k j w_j
+
O(1).
\end{align*}
Since $\sum_{j=1}^k w_j=1$, canceling the $\lambda_k\ N$ terms from both sides yields:
\[
\lambda_k
=
\frac{\sum_{j=1}^k w_j \expected{C_j}}{\sum_{j=1}^k j w_j}.
\]

Let $\mu_k \triangleq \sum_{j=1}^k j w_j$ denote the expected number of hops advanced in one speculative round. Using the definition of $w_j$, we obtain:
\[
\mu_k
=
\sum_{j=1}^{k-1} j p^{j-1}(1-p)+kp^{k-1}
=
\frac{1-p^k}{1-p}.
\]
Moreover, the expected number of speculative tool calls in a round is:
\[
\sum_{j=1}^k (j-1)w_j
=
\sum_{j=1}^k jw_j-\sum_{j=1}^k w_j
=
\mu_k-1.
\]

We now evaluate the numerator of $\lambda_k$ by substituting $\expected{C_j}$ and pulling the expectations out of the summations:
\begin{align*}
\sum_{j=1}^k w_j \expected{C_j}
&=
\sum_{j=1}^k w_j\left(j\expected{\Tseg} + (j-1)\expected{\Tspec} + \expected{\Ttool}\right) \\
&=
\expected{\Tseg}\sum_{j=1}^k jw_j
+
\expected{\Tspec}\sum_{j=1}^k (j-1)w_j
+
\expected{\Ttool}\sum_{j=1}^k w_j \\
&=
\expected{\Tseg}\mu_k + \expected{\Tspec}(\mu_k-1) + \expected{\Ttool}.
\end{align*}

Dividing by $\mu_k$ yields the expected latency per hop:
\[
\lambda_k
=
\frac{\expected{\Tseg}\mu_k + \expected{\Tspec}(\mu_k-1) + \expected{\Ttool}}{\mu_k}
=
\expected{\Tseg} + \frac{\mu_k-1}{\mu_k}\expected{\Tspec} + \frac{1}{\mu_k}\expected{\Ttool}.
\]

Substituting this exact rate back into our asymptotic assumption $\Latency_k(N)=\lambda_k N+O(1)$ gives the final expected runtime:
\[
\Latency_k(N)
=
N \left(
\expected{\Tseg} + \frac{\mu_k-1}{\mu_k}\expected{\Tspec} + \frac{1}{\mu_k}\expected{\Ttool}
\right)
+
O(1),
\]
where $\mu_k=\frac{1-p^k}{1-p}$.
This proves Theorem~\ref{thm:bounded-window-runtime}.

\subsection{More details on Pipeline Starvation}
\label{app:proof-starve}

\paragraph{Proof of Theorem \ref{thm:starvation-prob}}
Let the latencies $T_{\text{seg}}$, $T_{\text{spec}}$, and $T_{\text{target}}$ be independent random variables with means $\mu_{\text{seg}}, \mu_{\text{spec}}, \mu_{\text{target}}$ and variances $\sigma_{\text{seg}}^2, \sigma_{\text{spec}}^2, \sigma_{\text{target}}^2$. 

Pipeline starvation occurs when the  target tool outlasts the cumulative execution of $k$ active speculative threads. More specifically, starvation occurs if:
\[
T_{\text{target}} > \sum_{i=1}^k T_{\text{spec},i} + \sum_{i=2}^k T_{\text{seg},i}
\]

We define the time difference variable $D_k$, representing the latency buffer generated by the speculative pipeline:
\[
D_k = \sum_{i=1}^k T_{\text{spec},i} + \sum_{i=2}^k T_{\text{seg},i} - T_{\text{target}}
\]
Starvation occurs when $D_k < 0$. Assuming independence among the discrete execution steps, the expectation ($\mu_D$) and variance ($\sigma_D^2$) of $D_k$ are:
\[
\mu_D = k\mu_{\text{spec}} + (k-1)\mu_{\text{seg}} - \mu_{\text{target}}
\]
\[
\sigma_D^2 = k\sigma_{\text{spec}}^2 + (k-1)\sigma_{\text{seg}}^2 + \sigma_{\text{target}}^2
\]

By the Central Limit Theorem (CLT), for a sufficient sequence of steps, $D_k$ approximates a normal distribution $\mathcal{N}(\mu_D, \sigma_D^2)$. The probability of starvation is therefore evaluated using the standard normal cumulative distribution function $\Phi$:
\[
P(D_k < 0) = \Phi\left(\frac{0 - \mu_D}{\sigma_D}\right) = \Phi\left(\frac{\mu_{\text{target}} + (k-1)\mu_{\text{seg}} + \mu_{\text{seg}} - k\mu_{\text{spec}} - k\mu_{\text{seg}}}{\sigma_D}\right)
\]
Simplifying the numerator yields the generalized stochastic probability:
\[
P_{\text{starve}}(k) \approx \Phi\left(\frac{(\mu_{\text{target}} + \mu_{\text{seg}}) - k(\mu_{\text{spec}} + \mu_{\text{seg}})}{\sqrt{k\sigma_{\text{spec}}^2 + (k-1)\sigma_{\text{seg}}^2 + \sigma_{\text{target}}^2}}\right)
\]

To generalize this bound without relying on absolute wall-clock metrics, we leverage $\nu$, such that $\sigma_X \le \nu \mu_X$ for all system latencies. We substitute this bound into the denominator:
\[
\sigma_D \le \nu \sqrt{k\mu_{\text{spec}}^2 + (k-1)\mu_{\text{seg}}^2 + \mu_{\text{target}}^2}
\]

Next, we divide both the numerator and the variance components inside the square root by the target mean $\mu_{\text{target}}$, allowing us to substitute the dimensionless relative cost ratios
$\alpha = \mu_{\text{spec}}/\mu_{\text{target}}$ and $\beta = \mu_{\text{seg}}/\mu_{\text{target}}$:
\[
P_{\text{starve}}(k) \le \Phi\left(\frac{(1 + \beta) - k(\alpha + \beta)}{r \sqrt{k\alpha^2 + (k-1)\beta^2 + 1}}\right)
\]
This completes the proof, establishing a strict upper bound on starvation probability parameterized entirely by relative system costs and network volatility.

\begin{corollary}[Risk-Adjusted Capacity]
\label{cor:risk-adjusted}
To bound the starvation probability below an error rate $\epsilon$, the required active window $k^\star_{\epsilon}$ for \method{} is:
\[
k^\star_{\epsilon} \approx \left\lceil k_{\text{det}} + z_{1-\epsilon} \cdot \nu \cdot \frac{\sqrt{k_{\text{det}}\alpha^2 + (k_{\text{det}}-1)\beta^2 + 1}}{\alpha + \beta} \right\rceil
\]
where $z_{1-\epsilon}$ is the $(1-\epsilon)$-quantile of the standard normal distribution, and $k_{\text{det}} = \frac{1+\beta}{\alpha+\beta}$.
\end{corollary}

Theorem~\ref{thm:starvation-prob} shows that the required number of active threads depends on the relative latency profile of the system, captured by $\alpha$ and $\beta$.
Intuitively, when the speculator is much faster than the target tool, $\alpha$ is small, so the system can sustain a larger speculative window before verification returns.
In contrast, when model decoding dominates the hop latency, $\beta$ is larger, and fewer active threads are sufficient because the model cannot speculate far ahead before the target-tool result becomes available.

\textbf{Practical Examples ($\epsilon = 0.05$ for $95\%$ reliability).}
Based on our empirical results, system latencies exhibit a maximum volatility bound of $\nu \leq 0.4$.
We consider two representative settings.
First, in a fast-speculator and tool-intensive setting ($\alpha=0.2, \beta=0.15$), the deterministic baseline is $k_{\mathrm{det}} \approx 3.28$, and the stochastic bound requires $k^\star_{0.05}=6$ active threads.
Second, in a setting with a slower speculator and a less tool-intensive task ($\alpha=0.3, \beta=0.75$), the deterministic baseline is $k_{\mathrm{det}} \approx 1.67$, and only $k^\star_{0.05}=3$ active threads are required.
This suggests that, with high probability, near-optimal latency reduction can be achieved with a strictly limited amount of parallelism in realistic stochastic settings.

\section{Additional Results}
\label{app:add-exp}

\subsection{Live Execution}
\label{app:live}

We also run \method{} with the target-tool calls executing in parallel, at $k \rightarrow \infty$.
The E5 setting uses Llama~3.1~8B as $\speculator$ and 300 examples per dataset.
The web setting uses GPT-4o as $\speculator$ and 42--44 examples.
Table~\ref{tab:live-rellat} reports the measured wall-clock ratio $\RelLatencylive$ next to $\RelLatency$ from the procedure in Appendix~\ref{app:exp-setup} on the same examples.
On E5, a second $\generatormodel$ replica moves $\RelLatencylive$ from $0.950$, $0.903$, and $0.959$ to $0.925$, $0.880$, and $0.909$.

\input{tables/live_rellat}

Table~\ref{tab:live-contention} compares per-call times on the 2Wiki runs with the same stages timed in isolation.
On E5, $t_{\mathrm{seg}}$ rises from $216$\,ms to $227$\,ms and $t_{\mathrm{target}}$ from $426$\,ms to $457$\,ms.
With a second $\generatormodel$ replica those times are $181$\,ms and $376$\,ms.
On web search, $t_{\mathrm{seg}}$ stays at $179$\,ms and $t_{\mathrm{target}}$ rises from $10377$\,ms to $10739$\,ms.
On the E5 DeepResearch-9K run, the mean peak number of concurrent $\generatormodel$ calls is $2.2$ and of $\speculator$ calls is $1$.
On the web DeepResearch-9K run, the mean peak number of concurrent $\targettool$ calls is $6.6$.

\input{tables/live_contention}

\input{tables/gpt_baseline}

\subsection{\texorpdfstring{Generalization to State-of-the-Art Off-the-Shelf $\generatormodel$ (GPT-5)}{Generalization to State-of-the-Art Off-the-Shelf Generator (GPT-5)}}
\label{app:gpt_baseline}

While our primary experiments utilize CoRAG, we demonstrate the generalizability of \method{} by applying it to an off-the-shelf GPT-5 model. Rather than fine-tuning, we provide GPT-5 with a single-shot prompt (refer to Appendix~\ref{app:sys_prompts} for more details) demonstrating the desired multi-hop sub-query and answer format. We wrap this standard GPT-5 agent in our continuous speculation and verification framework, replacing its native sequential tool calls with our speculative threads.

Table~\ref{tab:gpt5-baseline} shows the relative latency achieved by our method, using GPT-4o mini as $\speculator$ and Web Search as $\targettool$. We can see that our method consistently achieves high relative latency in all settings, while preserving EM and F1 (Table~\ref{tab:app-em-f1-all}). One notable point about using GPT-5 as $\generatormodel$ is the higher generator latency, which can be observed directly in Figure~\ref{fig:latency-distributions} (top right plot), and higher $\beta$ for results in Table~\ref{tab:gpt5-baseline} compared to Table~\ref{tab:spec-metrics-three-datasets}.

\input{tables/em_f1_appendix}

\subsection{Effect of \method{} on EM/F1}
\label{app:em_f1_all}
While the primary advantage of \method{} is substantial wall-clock latency reduction, it is critical that these speedups do not compromise the task performance of the underlying retrieval-augmented trajectory. Because \method{} employs a strict verification step using the verifier $\verifier$ before committing to any prediction from the speculator model $\speculator$, the final trajectory effectively mirrors the one that would have been produced by the generator model $\generatormodel$ operating sequentially with the target tool $\targettool$. 

Table~\ref{tab:app-em-f1-all} details the Exact Match (EM) and F1 scores for both the standard sequential baseline and \method{} across all available experimental configurations, spanning three datasets, two target tools ($\targettool$), and multiple speculator models ($\speculator$). As shown, the task metrics remain nearly identical across all settings. Minor deviations (typically $<1\%$) arise strictly from benign variations accepted by the verifier $\verifier$ (e.g., functionally equivalent string variations or negligible API noise). This comprehensive breakdown confirms that \method{} successfully accelerates inference without degrading the reasoning or retrieval accuracy of either local generator models (CoRAG) or highly capable off-the-shelf generator models (GPT-5).

\input{figures/k_rellat_app}

\subsection{Human Evaluation of Verifier Performance}
\label{app:verifier-ablation}

\method{}'s losslessness depends on the verifier ($\verifier$) not accepting a divergent observation.
We labeled 500 verification decisions, 50 accepted and 50 rejected in each of five settings: 2WikiMultihopQA with Llama~3.1~8B and E5, and with GPT-4o and web search; MuSiQue with GPT-4o and web search, and with Llama~3.1~8B and web search; and DeepResearch-9K with GPT-4o and web search.
Table~\ref{tab:verifier-human} reports precision and recall against the human label.
Precision is $1.00$ on four of the five settings and $0.98$ on DeepResearch-9K with GPT-4o and web search. Overall precision is $0.996$ and recall is $0.91$.

Table~\ref{tab:verifier-length} splits the same labels by answer length.
On answers of at most 50 characters (474 of 500 hops), precision is $1.00$.
On the 26 longer answers, precision is $0.88$ and recall is $0.35$.
The default thresholds are $72\%$ token coverage or Jaccard similarity $0.55$.
Lowering coverage to $0.50$, with Jaccard held at $0.55$, moves precision from $0.996$ to $0.97$ and recall from $0.91$ to $0.93$.
Raising coverage to $0.90$ leaves recall at $0.91$ and raises precision to $1.00$.
Changing Jaccard from $0.55$ to $0.30$ or $0.80$, with coverage held at $0.72$, gives precision/recall $0.97/0.93$ and $0.996/0.91$.

\input{tables/verifier_human}
\input{tables/verifier_length}

\subsection{Speculative Actions Call Counts}
\label{app:spec-actions-calls}

Table~\ref{tab:spec-actions-calls} lists the call counts behind Table~\ref{tab:spec-actions}.
Ye et al.~\cite{ye2025speculative} do not keep speculating after the one-step lookahead, so the $\generatormodel$ count stays closer to the sequential baseline.

\input{tables/spec_actions_calls}

\subsection{Projected API Cost}
\label{app:cost}

Table~\ref{tab:api-cost} converts the DeepResearch-9K web-search call counts into a dollar figure using the rates in the caption.
The underlying run uses local CoRAG. The $\generatormodel$ and $\speculator$ columns are priced as GPT-4o and GPT-4o mini.

\input{tables/api_cost}

\input{tables/rellat_em-true}

\subsection{Latency Gain on Correct Predictions (EM=True)}
\label{app:em-true-latency}

The global latency gains reported in our main experiments encompass all evaluated samples, regardless of whether the generator model ($\generatormodel$) ultimately produced the correct final answer. As detailed in Table~\ref{tab:app-em-f1-all}, the intrinsic difficulty of the tasks varies significantly; for instance, MuSiQue is highly challenging for the baseline models (resulting in low Exact Match and F1 scores), whereas 2WikiMultihopQA is relatively easier (yielding high EM and F1 scores). 

In datasets with lower base performance, it is critical to ensure that the reported latency reductions are not simply an artifact of failing trajectories (e.g., the model getting confused or failing to retrieve useful context). To validate this, we filter the evaluation pipeline to isolate the subset of samples where the model successfully reasons through the problem and outputs the correct final answer (EM=True).

Table~\ref{tab:app-rellat-all} compares the overall relative latency ($\RelLatency$) with the relative latency on this successful subset ($\RelLatency_{EM=true}$). The results demonstrate that the lower overall accuracy on challenging datasets like MuSiQue does not invalidate \method{}'s efficiency gains. Across all datasets and speculator configurations, \method{} consistently achieves similar—and frequently even greater—latency reductions on the accurately predicted samples. This confirms that continuous speculation reliably accelerates successful, complex reasoning paths across all sample difficulties without bias toward failing trajectories.



\input{figures/distribution_metrics}

\subsection{Distribution of Latency Components}
\label{app:dists}

To better understand where time is spent during speculative execution, we plot the empirical distribution of stage runtimes, specifically the time spent on model generation ($t_{\mathrm{seg}}$), target tool execution ($t_{\mathrm{target}}$), and speculative execution ($t_{\mathrm{spec}}$). Figure~\ref{fig:latency-distributions} visualizes these distributions, providing critical insight into the system's temporal bottlenecks and the resulting efficiency gains. 

Looking at the individual stage latencies (top row), we observe a stark contrast between the target tools. The local E5 retriever is highly deterministic and fast ($\mu=0.44$s), whereas Web Search introduces a massive temporal bottleneck with a heavy right tail ($\mu=4.15$s, $\sigma=1.68$s) due to network-bound page fetching. The speculator latency ($t_{\mathrm{spec}}$) highlights the architectural divide between local and API-based models: local speculators like Llama 3.1 8B and Qwen 3 8B resolve in under 0.2 seconds, while API-dependent models like the GPT-4o series suffer from network overhead, pushing their mean execution times to roughly 0.6 seconds. Similarly, the generator latency ($t_{\mathrm{seg}}$) shows that while the local CoRAG model is tightly optimized ($\mu=0.33$s), querying the off-the-shelf GPT-5 model introduces significantly higher base latency ($\mu=1.48$s).

The bottom row translates these component latencies into system-wide speedups. Because Web Search dominates the sequential trajectory time, it offers the greatest opportunity for optimization. Consequently, the empirical relative latency ($\mathrm{RelLat}$) for Web Search tasks exhibits much lower medians and wider variance compared to E5 tasks, which naturally cluster closer to 1.0 due to their already low base latency. Finally, the scatter plot comparing Oracle to Empirical Relative Latency demonstrates that \method{} is highly efficient in practice; across all dataset and tool combinations, the empirical speedups closely hug the ideal $y=x$ line, proving that our continuous speculation framework successfully captures the vast majority of the theoretical maximum latency gain.



\section{Implementation Details}
\label{app:imp-details}

\subsection{Additional Implementation and Setup Details}
\label{app:exp-setup}

This section details the system configurations and execution rules omitted from the main text for brevity. 
All of our experiments are done with two NVIDIA L40 48GB GPUs, to load and use the generator model ($\generatormodel$), speculator ($\speculator$), and target tool ($\targettool$).

\paragraph{Generation Hyperparameters.} 
To ensure deterministic evaluation and minimize generation variance, all local generator models ($\generatormodel$) are served using vLLM with greedy decoding (\texttt{temperature = 0.0}). During standard multi-hop execution, the generator produces sub-queries with a 64-token cap, and intermediate/final answers with a 128-token cap. When utilizing LLMs as speculators ($\speculator$), we enforce a strict 16-token generation cap to minimize the speculative latency ($t_{\mathrm{spec}}$).

\paragraph{Dataset Splits and Early Stopping.} 
We evaluate on the \texttt{validation} splits for 2WikiMultihopQA and MuSiQue, and the \texttt{train} split for DeepResearch-9K (as it is the standard release split for the benchmark). During inference, to prevent infinite loops or degenerate trajectories, the pipeline halts if $\generatormodel$ generates the exact same sub-query twice (determined via normalized string matching). Furthermore, if the target tool ($\targettool$) repeatedly fails to find relevant context—returning the literal string ``No relevant information found''—we terminate the trajectory early. We allow one such failure for 2WikiMultihopQA and MuSiQue, and two failures for DeepResearch-9K before aborting.

\paragraph{Target Tool ($\targettool$) Backends.}
For the \textbf{E5 Retrieval} target tool, queries are encoded with a maximum length of 64 tokens using the \texttt{intfloat/e5-large-v2} model. For memory efficiency during parallel retrieval, the passage index is loaded via shards and queried using \texttt{int8} row-wise quantization. For the \textbf{Web Search} target tool, the system fetches the HTML of the returned DuckDuckGo URLs in parallel. Each page's visible text is extracted and truncated to 10,000 characters. The total target tool latency ($t_{\mathrm{target}}$) incorporates both the search API response time and the maximum page-fetch time across the parallel requests (capped strictly by the 3-second timeout).

\paragraph{Cached-Retrieval Speculator ($\speculator$).}
For the experiments evaluating a fast cache retriever as the speculator, the speculative backend is implemented as a smaller E5 index built from a heavily sharded subset of Wikipedia passages. The generator ($\generatormodel$) then produces the speculative observation ($\hat{\observation}_i$) using this truncated, rapidly retrieved context, following the same intermediate-answer prompt template as the target-tool path but strictly bounded by the shorter speculative token cap.

\paragraph{Latency Measurement and Simulation.}
To ensure strict comparability and isolate the temporal gains of \method{}, we execute the full set of required trajectory calls sequentially, recording the exact wall-clock times for every atomic stage: sub-query generation ($t_{\mathrm{seg}}$), target tool execution ($t_{\mathrm{target}}$), and speculative execution ($t_{\mathrm{spec}}$). The end-to-end relative latency ($\RelLatency$) for \method{} is then calculated using a deterministic discrete-event simulator that applies the continuous speculation schedule (Algorithm~\ref{alg:abstract-continuous-speculation}) over the empirically recorded stage times. This ensures that the reported speedups are unaffected by transient hardware loads or background system noise. Appendix~\ref{app:live} reports runs in which the target-tool calls execute in parallel.

\paragraph{Metrics.}
Answer quality (Exact Match and F1) is computed using standard SQuAD-style text normalization. In addition to measuring accuracy against the dataset's gold answers, we report the agreement between the speculative trajectory and the baseline trajectory to verify our theoretical losslessness.

\subsection{\texorpdfstring{System Prompts for $\generatormodel$}{System Prompts for the Generator}}
\label{app:sys_prompts}

For experiments utilizing CoRAG as the generator model ($\generatormodel$), we use the exact system prompts provided in the official implementation.\footnote{\url{https://github.com/microsoft/LMOps/tree/main/corag}}

For our off-the-shelf evaluation using GPT-5 as the generator model ($\generatormodel$), we design a structured, one-shot prompting pipeline to enforce the required multi-hop reasoning and tool-use trajectory. The pipeline consists of a universal system instruction and step-specific user prompts for each stage of the generation process. To ensure strict structural compliance and easy parsing by the verifier ($\verifier$), we utilize XML-style tags.

\subsubsection*{Universal System Instruction}
This system prompt establishes the strict behavioral constraints and provides a single-shot example to align the off-the-shelf model with our trajectory format.

\begin{mdframed}[backgroundcolor=gray!10, linewidth=0.5pt, roundcorner=4pt, innertopmargin=10pt, innerbottommargin=10pt]
\small\ttfamily
You are a strict multi-hop QA agent. Follow the required XML-like tags exactly.

Rules:
1) Only output one of the allowed tags at a time based on the current instruction.
2) Allowed tags: \textless subquestion\textgreater...\textless/subquestion\textgreater, \textless subanswer\textgreater...\textless/subanswer\textgreater, \textless final\_answer\textgreater...\textless/final\_answer\textgreater.
3) Never output explanations, markdown, JSON, or any text outside the required tag.
4) Keep subquestions simple and answerable via web search snippets.
5) Subquestions should target a single short factual answer (name, year, place, yes/no).
6) Subanswers should be a single short factual answer (name, year, place, yes/no).
7) Use at most \{max\_hops\} subquestions, but stop as soon as the final answer can be derived from the subanswers with certainty.

Format example to imitate:
question: "Which film has the director who was born earlier, Face Of A Fugitive or Cage Of Gold?"
output: 
\textless subquestion\textgreater\ Who directed Face Of A Fugitive? \textless/subquestion\textgreater
\textless subanswer\textgreater\ Paul Wendkos \textless/subanswer\textgreater
\textless subquestion\textgreater\ What is the birth year of Paul Wendkos? \textless/subquestion\textgreater
\textless subanswer\textgreater\ 1925 \textless/subanswer\textgreater
\textless subquestion\textgreater\ Who directed Cage Of Gold \textless/subquestion\textgreater
\textless subanswer\textgreater\ Basil Dearden \textless/subanswer\textgreater
\textless subquestion\textgreater\ What is the birth year of Basil Dearden? \textless/subquestion\textgreater
\textless subanswer\textgreater\ 1911 \textless/subanswer\textgreater

\textless final\_answer\textgreater\ Cage Of Gold \textless/final\_answer\textgreater
\end{mdframed}

\subsubsection*{Step-Specific User Prompts}

At each step of the trajectory, the generator model ($\generatormodel$) receives the universal system prompt along with one of the following specific instructions, depending on the current state of the inference loop.

\paragraph{1. Next Sub-question Generation.} Invoked when the model needs to query the target tool ($\targettool$) or speculator ($\speculator$).
\begin{mdframed}[backgroundcolor=gray!5, linewidth=0.5pt, roundcorner=4pt]
\small\ttfamily
Main question:\\
\{question\}

Current hop index: \{hops\_done + 1\} / \{max\_hops\}\\
Existing reasoning trace:\\
\{trace\_blob\}

Output exactly one next subquestion in this format:\\
\textless subquestion\textgreater\ ... \textless/subquestion\textgreater
\end{mdframed}

\paragraph{2. Sub-answer Generation.} Invoked to process the retrieved documents from the tool.
\begin{mdframed}[backgroundcolor=gray!5, linewidth=0.5pt, roundcorner=4pt]
\small\ttfamily
Main question:\\
\{question\}

Current subquestion:\\
\{subquestion\}

Search results:\\
\{search\_snippets\}

Answer the current subquestion only using the search results. If uncertain, return a short best-effort answer.\\
Output exactly:\\
\textless subanswer\textgreater\ ... \textless/subanswer\textgreater
\end{mdframed}

\paragraph{3. Final Answer Generation.} Invoked when the hop budget is exhausted or sufficient information is gathered.
\begin{mdframed}[backgroundcolor=gray!5, linewidth=0.5pt, roundcorner=4pt]
\small\ttfamily
Main question:\\
\{question\}

Reasoning trace:\\
\{trace\_blob\}

Output exactly one final answer tag:\\
\textless final\_answer\textgreater\ ... \textless/final\_answer\textgreater
\end{mdframed}

\paragraph{Formatting Fallback Mechanism.} 
In the rare event that the off-the-shelf model outputs reasoning tokens outside the designated XML tags or deviates from the schema, we use a lightweight fallback prompt to force structural compliance before advancing the trajectory:
\begin{mdframed}[backgroundcolor=gray!5, linewidth=0.5pt, roundcorner=4pt]
\small\ttfamily
Your previous output violated the required format.\\
Previous output:\\
\{bad\_output\}

Return ONLY one valid \{required\_tag\} tag and nothing else.
\end{mdframed}

\subsection{\texorpdfstring{System Prompts for $\speculator$}{System Prompts for the Speculator}}

For all LLM-based speculator models ($\speculator$), we employ a highly concise, few-shot prompting strategy designed to minimize generation latency and enforce strictly formatted, direct answers. The speculator is explicitly instructed to act as a ``fast assistant'' and bypass any conversational filler or step-by-step reasoning. 

\subsubsection*{Universal System Instruction}
This system prompt establishes the rapid-response persona and provides four diverse few-shot examples to demonstrate the expected question-answer mapping using XML-style tags.

\begin{mdframed}[backgroundcolor=gray!10, linewidth=0.5pt, roundcorner=4pt, innertopmargin=10pt, innerbottommargin=10pt]
\small\ttfamily
You are a fast assistant. Return only the direct answer, with no explanation.

\textless question\textgreater What is the name of the performer of the song Velvet Goldmine?\textless/question\textgreater\\
\textless answer\textgreater T. Rex\textless/answer\textgreater

\textless question\textgreater Where is Babu Nanthankode from?\textless/question\textgreater\\
\textless answer\textgreater Thiruvananthapuram, India.\textless/answer\textgreater

\textless question\textgreater What is the birthdate of Hans Olsen (Cyclist)?\textless/question\textgreater\\
\textless answer\textgreater 1 January 1885\textless/answer\textgreater

\textless question\textgreater Is Reservoir High School located in Victoria?\textless/question\textgreater\\
\textless answer\textgreater Yes.\textless/answer\textgreater
\end{mdframed}

\subsubsection*{Sub-query Speculation Prompt}
At inference time, the specific sub-query generated by $\generatormodel$ is injected into the user prompt. We append the opening \texttt{\textless answer\textgreater} tag to force the speculator model to immediately begin generating the factual response without preamble.

\begin{mdframed}[backgroundcolor=gray!5, linewidth=0.5pt, roundcorner=4pt]
\small\ttfamily
\textless question\textgreater \{subquery\}\textless/question\textgreater\\
\textless answer\textgreater
\end{mdframed}

\paragraph{Generation Parameters and Post-Processing.} 
To further ensure optimal latency and deterministic behavior, all speculator calls are executed using greedy decoding (\texttt{temperature = 0.0}) and a strict length limit (\texttt{max\_new\_tokens = 16}). After generation, a lightweight post-processing step automatically strips any residual \texttt{\textless answer\textgreater} or \texttt{\textless/answer\textgreater} tags before passing the predicted observation to the verifier ($\verifier$).

\subsection{\texorpdfstring{Deterministic Rule-Based Verifier ($\verifier$) Implementation}{Deterministic Rule-Based Verifier Implementation}}
\label{app:verifier-details}

To ensure trajectory losslessness without the latency overhead of an LLM-based evaluator, we utilize a lightweight, deterministic rule-based verifier ($\verifier$). Because the integrity of the speculative framework relies entirely on preventing false positives, our implementation employs a set of conservative heuristics to guarantee that only factually equivalent observations are accepted.

The verifier compares the speculative observation ($\hat{\observation}_i$) with the target tool's observation ($\observation_i$) through a strict evaluation pipeline. First, both texts undergo standard normalization (lowercasing, diacritic and punctuation removal). The verifier then immediately rejects any speculative outputs that match common refusal or uncertainty patterns (e.g., ``I don't know,'' ``information unavailable'') to prevent the trajectory from absorbing uninformative responses. 

For factual verification, the system enforces a strict numeric consistency check: if the target observation contains multi-digit numbers (such as years or quantities), the speculative observation must contain matching numeric values, otherwise it is automatically rejected. Final acceptance requires either a direct substring match or a high degree of lexical overlap. Specifically, after filtering out standard English stopwords, the speculation must achieve either $\ge 72\%$ token coverage or a Jaccard similarity of $\ge 0.55$. For exceptionally short answers (under 5 characters), the verifier defaults to requiring a perfect token match. This robust, cautious design ensures that speculative threads only commit when they deliver the precise factual payload required by the generator model ($\generatormodel$).


%% file: tables/live_rellat.tex
\begin{table}[t]
  \centering
  \small
  \renewcommand{\arraystretch}{1.2}
  \setlength{\tabcolsep}{4pt}
  \caption{Parallel target-tool execution of \method{} at $k \rightarrow \infty$. $\RelLatencylive$ is measured wall-clock over sequential execution on the same examples. $\RelLatency$ and $\RelLatency^*$ use the latency calculation in Appendix~\ref{app:exp-setup}.}
  \vspace{2pt}
  \begin{tabular}{@{}l c c c c@{}}
    \toprule
    Setting & $\hat{p}$ & $\RelLatencylive$ $\downarrow$ & $\RelLatency$ $\downarrow$ & $\RelLatency^*$ $\downarrow$ \\
    \midrule
    E5, Llama~3.1~8B, 2Wiki & 0.258 & 0.950 & 0.899 & 0.866 \\
    E5, Llama~3.1~8B, MuSiQue & 0.353 & 0.903 & 0.859 & 0.807 \\
    E5, Llama~3.1~8B, DeepRes. & 0.261 & 0.959 & 0.900 & 0.893 \\
    \midrule
    Web, GPT-4o, 2Wiki & 0.532 & 0.718 & 0.749 & 0.506 \\
    Web, GPT-4o, MuSiQue & 0.546 & 0.732 & 0.719 & 0.501 \\
    Web, GPT-4o, DeepRes. & 0.423 & 0.783 & 0.712 & 0.612 \\
    \bottomrule
  \end{tabular}
  \label{tab:live-rellat}
\end{table}

%% file: tables/live_contention.tex
\begin{table}[t]
  \centering
  \small
  \renewcommand{\arraystretch}{1.15}
  \setlength{\tabcolsep}{4pt}
  \caption{Mean per-call time (ms) on the 2Wiki runs in Appendix~\ref{app:live}. Parallel is under concurrent target-tool calls. Isolated is the same stage timed alone.}
  \label{tab:live-contention}
  \vspace{2pt}
  \begin{tabular}{@{}l cccc@{}}
    \toprule
    Setting & $t_{\mathrm{seg}}$ parallel & $t_{\mathrm{seg}}$ isolated & $t_{\mathrm{target}}$ parallel & $t_{\mathrm{target}}$ isolated \\
    \midrule
    E5, Llama 3.1 8B & 227 & 216 & 457 & 426 \\
    Web, GPT-4o & 179 & 179 & 10739 & 10377 \\
    \bottomrule
  \end{tabular}
\end{table}

%% file: tables/gpt_baseline.tex

\definecolor{metricbg}{HTML}{F4F7F9} 
\definecolor{bestval}{HTML}{0A3B5C}  

\begin{table*}[t]
  \centering
  \renewcommand{\arraystretch}{1.5}
  \setlength{\tabcolsep}{3pt} 
  
  \resizebox{\textwidth}{!}{%
  \begin{tabular}{@{}c l ccc >{\columncolor{metricbg}}c >{\columncolor{metricbg}}c ccc >{\columncolor{metricbg}}c >{\columncolor{metricbg}}c ccc >{\columncolor{metricbg}}c >{\columncolor{metricbg}}c@{}}
    \toprule
    & & \multicolumn{5}{c}{\textbf{2WikiMultihopQA}} 
    & \multicolumn{5}{c}{\textbf{MuSiQue}} 
    & \multicolumn{5}{c}{\textbf{DeepResearch-9K}} \\
    
    \cmidrule(lr){3-7} \cmidrule(lr){8-12} \cmidrule(lr){13-17}
    
    & Speculator ($\mathcal{S}$)
      & $p$ & $\alpha$ & $\beta$ & $\RelLatency^*$ $\downarrow$ & $\RelLatency$ $\ \downarrow$
      & $p$ & $\alpha$ & $\beta$ & $\RelLatency^*$ $\downarrow$ & $\RelLatency$ $\ \downarrow$
      & $p$ & $\alpha$ & $\beta$ & $\RelLatency^*$ $\downarrow$ & $\RelLatency$ $\ \downarrow$ \\
    \midrule \midrule
    
    \multirow{1}{*}{\textbf{Web}}
      & GPT-4o mini
        & 0.44 & 0.15 & 0.34 & 0.72 & 0.78
        & 0.56 & 0.14 & 0.23 & 0.61 & 0.70
        & 0.67 & 0.15 & 0.26 & 0.55 & 0.60 \\
    \bottomrule
  \end{tabular}%
  }
    \caption{Latency reduction achieved by \method{} using an off-the-shelf \textbf{GPT-5} as the generator model ($\generatormodel$). The target tool is Web Search, and the speculator is GPT-4o mini. The results demonstrate that even when wrapping a highly capable, generalized reasoning model, continuous speculation yields significant wall-clock speedups across all multi-hop datasets.}
    \label{tab:gpt5-baseline}
\end{table*}


  

%% file: tables/em_f1_appendix.tex
\begin{table*}[t]
  \centering
  \renewcommand{\arraystretch}{1.3}
  \setlength{\tabcolsep}{4pt} 
  
  \resizebox{\textwidth}{!}{%
  \begin{tabular}{@{}c l cc >{\columncolor{metricbg}}c >{\columncolor{metricbg}}c cc >{\columncolor{metricbg}}c >{\columncolor{metricbg}}c cc >{\columncolor{metricbg}}c >{\columncolor{metricbg}}c@{}}
    \toprule
    & & \multicolumn{4}{c}{\textbf{2WikiMultihopQA}} 
    & \multicolumn{4}{c}{\textbf{MuSiQue}} 
    & \multicolumn{4}{c}{\textbf{DeepResearch-9K}} \\
    
    \cmidrule(lr){3-6} \cmidrule(lr){7-10} \cmidrule(lr){11-14}
    
    & & \multicolumn{2}{c}{Standard} & \multicolumn{2}{c}{\method{}}
      & \multicolumn{2}{c}{Standard} & \multicolumn{2}{c}{\method{}}
      & \multicolumn{2}{c}{Standard} & \multicolumn{2}{c}{\method{}} \\
    \cmidrule(lr){3-4} \cmidrule(lr){5-6} 
    \cmidrule(lr){7-8} \cmidrule(lr){9-10} 
    \cmidrule(lr){11-12} \cmidrule(lr){13-14}
    
    & Speculator Model ($\speculator$)
      & EM & F1 & EM & F1 
      & EM & F1 & EM & F1 
      & EM & F1 & EM & F1 \\
    \midrule \midrule
    
    \multirow{3}{*}{\textbf{E5}}
      & Llama 3.1 8B
        & 72.3 & 76.9 & 72.3 & 76.9
        & 23.7 & 36.5 & 23.7 & 36.6
        & 22.0 & 30.1 & 22.3 & 30.3 \\
      & Qwen 3 8B
        & 72.3 & 76.7 & 72.3 & 76.7
        & 23.3 & 36.6 & 23.7 & 37.0
        & 26.0 & 34.1 & 26.3 & 34.2 \\
      & GPT-4o mini
        & 73.3 & 80.7 & 73.3 & 80.7
        & 23.3 & 36.6 & 23.3 & 36.8
        & 27.0 & 34.6 & 26.7 & 34.6 \\
    \midrule
    
    \multirow{4}{*}{\textbf{Web}}
      & Llama 3.1 8B
        & 68.7 & 73.8 & 69.3 & 73.8
        & 10.0 & 25.1 & 10.7 & 25.3
        & 32.0 & 35.6 & 32.0 & 35.6 \\
      & Qwen 3 8B
        & 67.3 & 71.1 & 67.3 & 71.1
        & 10.0 & 23.3 & 10.0 & 23.2
        & 26.7 & 32.4 & 26.7 & 32.4 \\
      & GPT-4o mini
        & 65.3 & 70.7 & 66.0 & 70.7
        & 14.0 & 26.9 & 15.3 & 27.3
        & 28.7 & 32.3 & 28.0 & 32.0 \\
      & GPT-4o
        & 69.3 & 74.2 & 70.0 & 74.5
        & 10.7 & 24.4 & 10.7 & 24.4
        & 26.0 & 31.3 & 26.0 & 31.3 \\
      & GPT-4o mini ($\generatormodel$: GPT-5)
        & 66.0 & 77.1 & 64.7 & 76.2
        & 20.0 & 37.6 & 20.7 & 37.8
        & 30.0 & 40.2 & 30.7 & 40.6 \\
    \bottomrule
  \end{tabular}%
  }
  \caption{Comprehensive breakdown of Exact Match (EM) and F1 scores across all evaluated configurations. \method{} tightly preserves the task accuracy of the standard sequential execution. Unless otherwise noted, the generator model ($\generatormodel$) is CoRAG. The final row reflects experiments where the standard trajectory was produced using GPT-5 as the generator model ($\generatormodel$), maintaining accuracy even with a highly capable off-the-shelf model.}
  \label{tab:app-em-f1-all}
\end{table*}

%% file: figures/k_rellat_app.tex
\begin{figure}[htbp]
    \centering
    
    \textbf{$\generatormodel$: CoRAG, $\speculator$: Llama 3.1 8B, $\targettool$: E5 Retrieval}\\[0.5em]
    \includegraphics[width=0.95\linewidth]{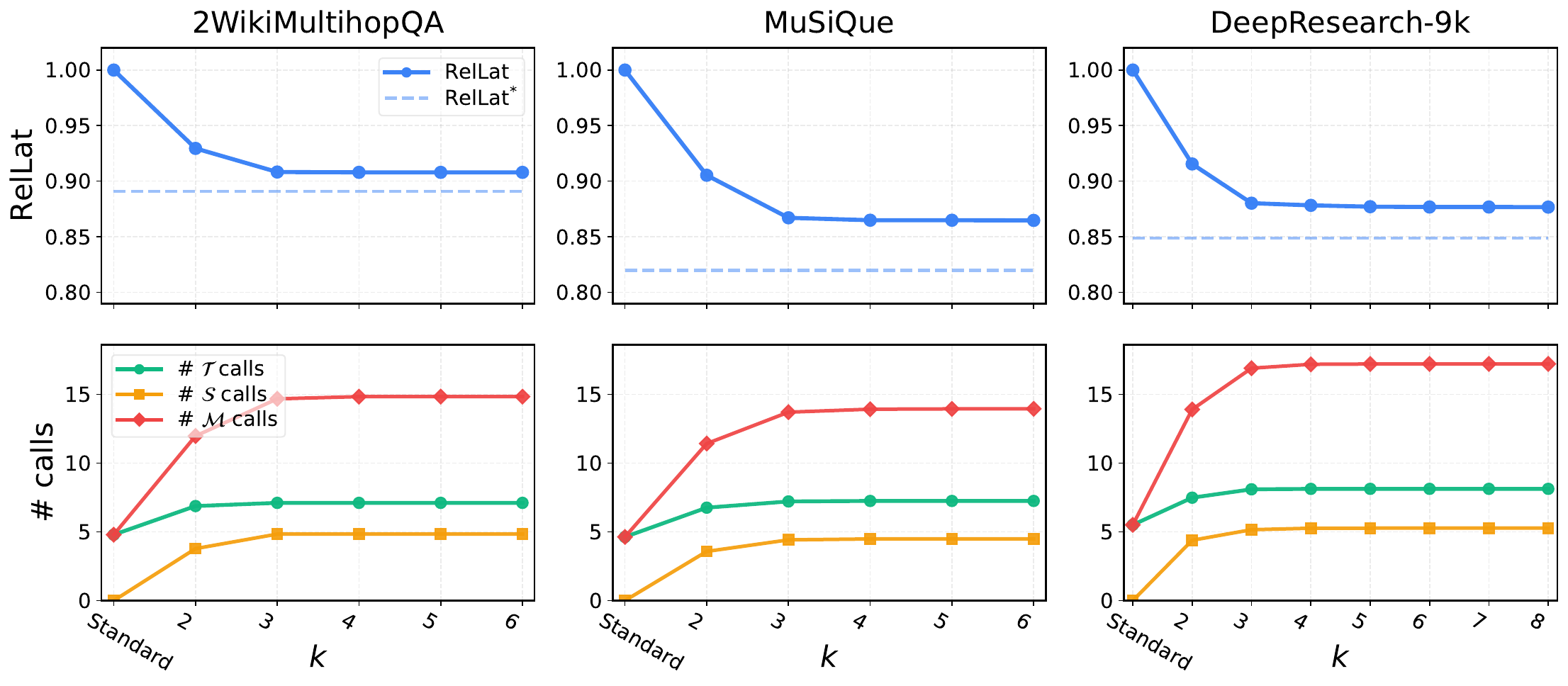}
    
    \vspace{1.05em}
    
    \textbf{$\generatormodel$: CoRAG, $\speculator$: Llama 3.1 8B, $\targettool$: Web Search}\\[0.5em]
    \includegraphics[width=0.95\linewidth]{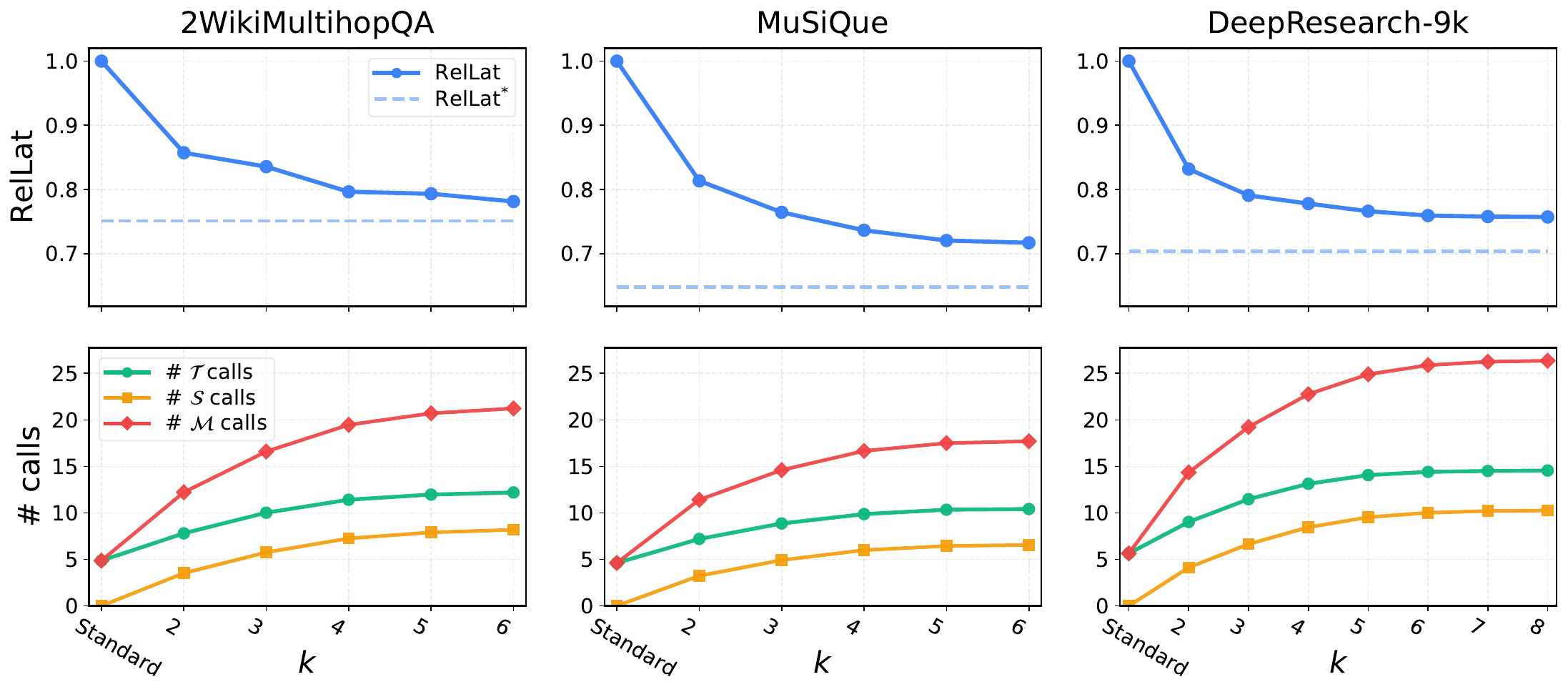}
    
    \vspace{1.05em}
    
    \textbf{$\generatormodel$: GPT-5, $\speculator$: GPT-4o mini, $\targettool$: Web Search}\\[0.5em]
    \includegraphics[width=0.95\linewidth]{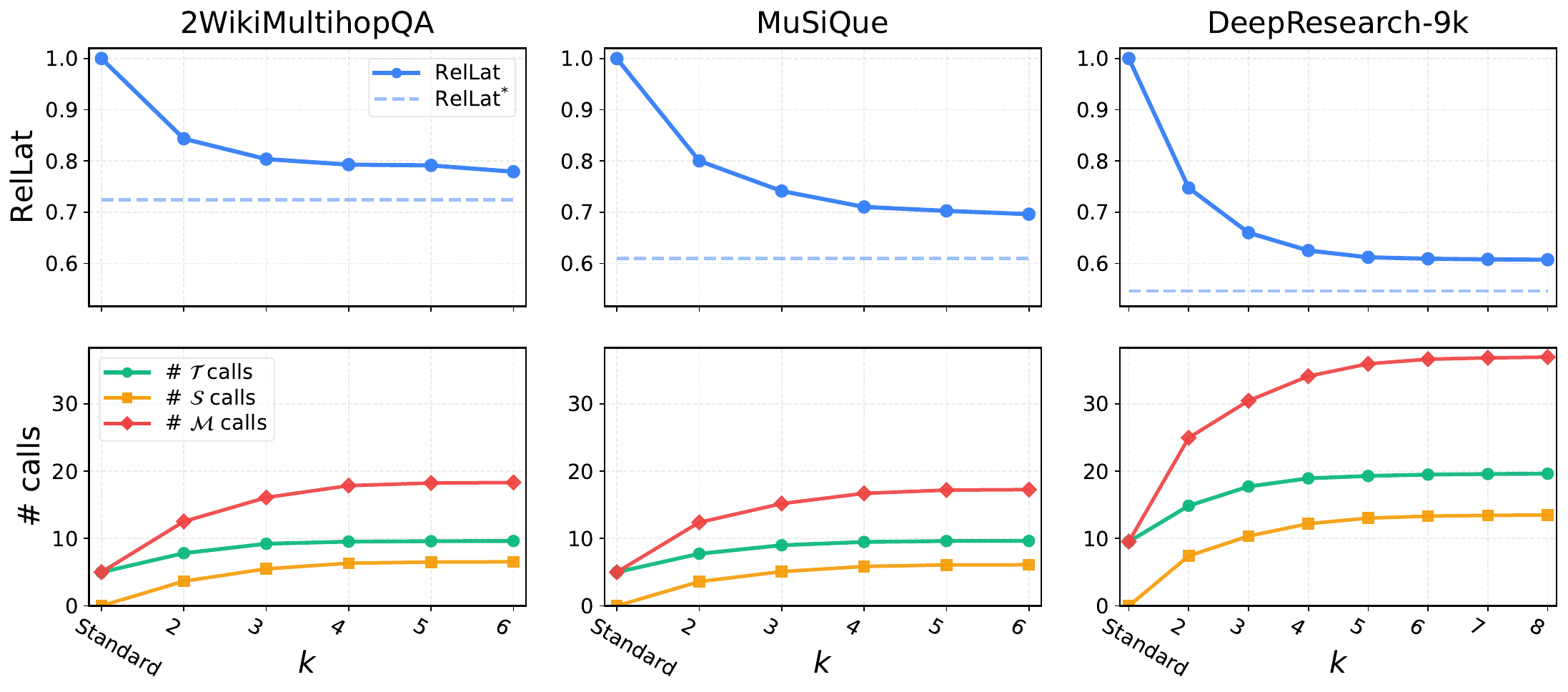}
    
    \caption{The effect of the active thread limit ($k$) on relative latency ($\RelLatency$) and computational cost across various extended configurations. The plots track the average number of calls to the target tool ($\targettool$), speculator ($\speculator$), and generator model ($\generatormodel$). The first two rows utilize CoRAG as the standard generator model, exploring different combinations of speculators and target tools. The final row demonstrates the $k$-capacity trade-off when wrapping a highly capable off-the-shelf GPT-5 model (refer to Appendix~\ref{app:gpt_baseline} for more details).}
    \label{fig:k_rellat_appendix_all}
\end{figure}

%% file: tables/verifier_human.tex
\begin{table}[t]
  \centering
  \small
  \renewcommand{\arraystretch}{1.15}
  \setlength{\tabcolsep}{4pt}
  \caption{Blind human evaluation of the verifier, 50 accepted and 50 rejected decisions per row (500 total). Precision and recall treat the human label as ground truth.}
  \label{tab:verifier-human}
  \vspace{2pt}
  \begin{tabular}{@{}l l cc@{}}
    \toprule
    Dataset & $\speculator$ / $\targettool$ & Prec. & Rec. \\
    \midrule
    2WikiMultihopQA & Llama 3.1 8B / E5 & 1.00 & 0.98 \\
    2WikiMultihopQA & GPT-4o / Web & 1.00 & 0.88 \\
    MuSiQue & GPT-4o / Web & 1.00 & 0.91 \\
    DeepResearch-9K & GPT-4o / Web & 0.98 & 0.93 \\
    MuSiQue & Llama 3.1 8B / Web & 1.00 & 0.89 \\
    \midrule
    Overall & 500 samples & 0.996 & 0.91 \\
    \bottomrule
  \end{tabular}
\end{table}

%% file: tables/verifier_length.tex
\begin{table}[t]
  \centering
  \small
  \renewcommand{\arraystretch}{1.15}
  \setlength{\tabcolsep}{5pt}
  \caption{Verifier precision and recall by answer length on the 500 human labels in Table~\ref{tab:verifier-human}.}
  \label{tab:verifier-length}
  \vspace{2pt}
  \begin{tabular}{@{}l ccc@{}}
    \toprule
    Answer length & $N$ & Prec. & Rec. \\
    \midrule
    $\le 5$ chars & 151 & 1.00 & 0.96 \\
    6--20 chars & 257 & 1.00 & 0.96 \\
    21--50 chars & 66 & 1.00 & 0.91 \\
    $> 50$ chars & 26 & 0.88 & 0.35 \\
    \bottomrule
  \end{tabular}
\end{table}

%% file: tables/spec_actions_calls.tex
\begin{table}[t]
  \centering
  \small
  \renewcommand{\arraystretch}{1.1}
  \setlength{\tabcolsep}{4pt}
  \caption{Average target-tool ($\targettool$), speculator ($\speculator$), and generator ($\generatormodel$) calls for the comparison in Table~\ref{tab:spec-actions}.}
  \vspace{2pt}
  \begin{tabular}{@{}ll ccc@{}}
    \toprule
    Dataset & Method & $\targettool$ & $\speculator$ & $\generatormodel$ \\
    \midrule
    \multirow{5}{*}{2Wiki}
      & Standard & 4.87 & 0.00 & 4.87 \\
      & Ye et al.~\cite{ye2025speculative} & 5.72 & 2.61 & 7.48 \\
      & \method{}, $k=2$ & 7.31 & 3.37 & 11.89 \\
      & \method{}, $k=3$ & 8.29 & 4.42 & 13.81 \\
      & \method{}, $k \rightarrow \infty$ & 9.06 & 5.31 & 15.40 \\
    \midrule
    \multirow{5}{*}{MuSiQue}
      & Standard & 4.60 & 0.00 & 4.60 \\
      & Ye et al.~\cite{ye2025speculative} & 5.96 & 2.61 & 7.21 \\
      & \method{}, $k=2$ & 7.11 & 3.23 & 11.24 \\
      & \method{}, $k=3$ & 8.49 & 4.61 & 13.85 \\
      & \method{}, $k \rightarrow \infty$ & 9.28 & 5.75 & 15.88 \\
    \midrule
    \multirow{5}{*}{DeepResearch-9K}
      & Standard & 5.48 & 0.00 & 5.48 \\
      & Ye et al.~\cite{ye2025speculative} & 7.53 & 3.37 & 8.85 \\
      & \method{}, $k=2$ & 8.67 & 4.01 & 13.76 \\
      & \method{}, $k=3$ & 10.69 & 6.14 & 17.83 \\
      & \method{}, $k \rightarrow \infty$ & 12.07 & 8.22 & 21.65 \\
    \bottomrule
  \end{tabular}
  \label{tab:spec-actions-calls}
\end{table}

%% file: tables/api_cost.tex
\begin{table}[t]
  \centering
  \small
  \renewcommand{\arraystretch}{1.15}
  \setlength{\tabcolsep}{4pt}
  \caption{DeepResearch-9K call counts from the web-search $k$ sweep (GPT-4o as $\speculator$), with a projected API cost. Web search is priced at \$0.005 per request. $\speculator$ is priced as GPT-4o mini (\$0.15 / \$0.60 per 1M input/output tokens) and $\generatormodel$ as GPT-4o (\$2.50 / \$10.00), which gives about \$0.0003 and \$0.0060 per call. The sweep itself uses local CoRAG; the $\generatormodel$ and $\speculator$ prices are a projection onto those APIs.}
  \label{tab:api-cost}
  \vspace{2pt}
  \begin{tabular}{@{}c c ccc c@{}}
    \toprule
    $k$ & $\RelLatency$ & $\targettool$ & $\speculator$ & $\generatormodel$ & \$ / query \\
    \midrule
    1 & 1.00 & 5.48 & 0.00 & 5.48 & 0.06 \\
    2 & 0.81 & 8.67 & 4.01 & 13.76 & 0.13 \\
    3 & 0.75 & 10.69 & 6.14 & 17.83 & 0.16 \\
    4 & 0.73 & 11.66 & 7.53 & 20.40 & 0.18 \\
    5 & 0.72 & 12.01 & 8.08 & 21.41 & 0.19 \\
    10 & 0.71 & 12.07 & 8.22 & 21.65 & 0.19 \\
    \bottomrule
  \end{tabular}
\end{table}

%% file: tables/rellat_em-true.tex
\begin{table*}[t]
  \centering
  \renewcommand{\arraystretch}{1.3}
  \setlength{\tabcolsep}{6pt}

  \resizebox{\textwidth}{!}{%
  \begin{tabular}{@{}c l c >{\columncolor{metricbg}}c c >{\columncolor{metricbg}}c c >{\columncolor{metricbg}}c@{}}
    \toprule
    & & \multicolumn{2}{c}{\textbf{2WikiMultihopQA}}
    & \multicolumn{2}{c}{\textbf{MuSiQue}}
    & \multicolumn{2}{c}{\textbf{DeepResearch-9K}} \\

    \cmidrule(lr){3-4} \cmidrule(lr){5-6} \cmidrule(lr){7-8}

    & Speculator Model ($\speculator$)
      & $\RelLatency$ & $\RelLatency_{EM=true}$ 
      & $\RelLatency$  & $\RelLatency_{EM=true}$  
      & $\RelLatency$  & $\RelLatency_{EM=true}$   \\
    \midrule \midrule

    \multirow{3}{*}{\textbf{E5}}
      & Llama 3.1 8B
        & 0.91 & 0.91
        & 0.87 & 0.86
        & 0.88 & 0.84 \\
      & Qwen 3 8B
        & 0.95 & 0.96
        & 0.89 & 0.91
        & 0.92 & 0.87 \\
    \midrule

    \multirow{5}{*}{\textbf{Web}}
      & Llama 3.1 8B
        & 0.78 & 0.77
        & 0.72 & 0.71
        & 0.76 & 0.65 \\
      & Qwen 3 8B
        & 0.88 & 0.87
        & 0.78 & 0.75
        & 0.78 & 0.65 \\
      & GPT-4o mini
        & 0.76 & 0.76
        & 0.72 & 0.76
        & 0.75 & 0.63 \\
      & GPT-4o
        & 0.60 & 0.59
        & 0.69 & 0.64
        & 0.72 & 0.60 \\
      & GPT-4o mini ($\generatormodel$: GPT-5)
        & 0.78 & 0.76
        & 0.70 & 0.66
        & 0.60 & 0.56 \\
    \bottomrule
  \end{tabular}%
  }
  \caption{Comparison of overall empirical relative latency ($\RelLatency$) against the relative latency isolated to the subset of queries answered correctly by the base model ($\RelLatency_{EM=true}$). \method{} consistently delivers robust speedups on successful multi-hop trajectories, confirming that latency reductions hold steady regardless of the underlying sample difficulty (e.g., the challenging MuSiQue versus the easier 2WikiMultihopQA).}
  \label{tab:app-rellat-all}
\end{table*}

%% file: figures/distribution_metrics.tex
\begin{figure*}[t]
    \centering
    \includegraphics[width=\textwidth]{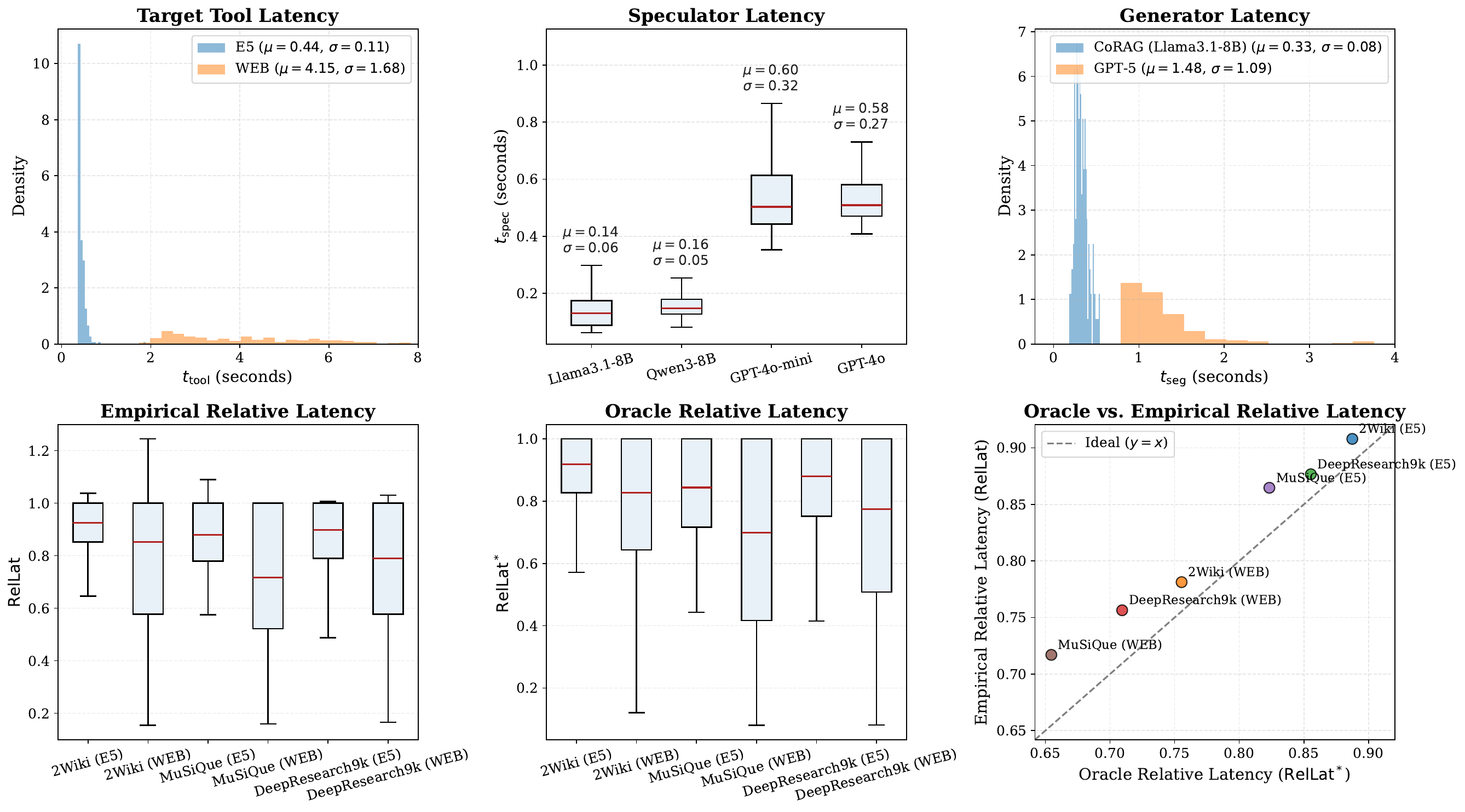} 
    \caption{Empirical distributions of hop-level latency components and resulting relative latency metrics. Mean ($\mu$) and standard deviation ($\sigma$) are annotated for temporal components. Unless otherwise specified on the axes, the default configuration uses CoRAG as the generator model ($\generatormodel$), Llama 3.1 8B as the speculator ($\speculator$), 2WikiMultihopQA as the dataset, and Web Search as target tool ($\targettool$). The top row isolates the variance inherent to the target tool ($t_{\mathrm{tool}}$), the speculator ($t_{\mathrm{spec}}$), and the generator model ($t_{\mathrm{seg}}$), including a direct comparison between local CoRAG and API-based GPT-5 generation speeds. The bottom row highlights the practical speedups ($\mathrm{RelLat}$) achieved across different datasets and target tools compared to their theoretical limits ($\mathrm{RelLat}^*$).}
    \label{fig:latency-distributions}
\end{figure*}